\def\cvprPaperID{10347}
\def\confName{ICCV}
\def\confYear{2023}

\def\authorBlock{
    Ziwei Liu\\
    Peking University\\
    Beijing, China\\
    {\tt\small liuziwei@stu.pku.edu.cn}
    \and
    Yongtao Wang\thanks{Corresponding author.}\\
    Peking University\\
    Beijing, China\\
    {\tt\small wyt@pku.edu.cn}
    \and
    Xiaojie Chu \\
    Peking University\\
    Beijing, China\\
    {\tt\small chuxiaojie@stu.pku.edu.cn}
}

\newif\ifreview 
\newif\ifarxiv \newcommand{\arxiv}{\arxivtrue}
\newif\ifcamera 
\newif\ifrebuttal 

\arxiv % \review OR \arxiv OR \cameraready

\documentclass[10pt,twocolumn,letterpaper]{article}
\ifreview \usepackage[review]{cvpr} \fi
\ifarxiv \usepackage[pagenumbers]{cvpr} \fi
\ifrebuttal \usepackage[rebuttal]{cvpr} \fi
\ifcamera \usepackage{cvpr} \fi
\usepackage{float}

\usepackage{graphicx}
\usepackage{amsmath}
\usepackage{amssymb}
\usepackage{booktabs}
\usepackage{pifont}
\usepackage{algorithm}
\usepackage{listings}

\usepackage{etoolbox}
\makeatletter
\AfterEndEnvironment{algorithm}{\let\@algcomment\relax}
\AtEndEnvironment{algorithm}{\kern2pt\hrule\relax\vskip3pt\@algcomment}
\let\@algcomment\relax
\newcommand\algcomment[1]{\def\@algcomment{\footnotesize#1}}
\renewcommand\fs@ruled{\def\@fs@cfont{\bfseries}\let\@fs@capt\floatc@ruled
  \def\@fs@pre{\hrule height.8pt depth0pt \kern2pt}%
  \def\@fs@post{}%
  \def\@fs@mid{\kern2pt\hrule\kern2pt}%
  \let\@fs@iftopcapt\iftrue}
\makeatother
\usepackage{times}
\usepackage{microtype}
\usepackage{epsfig}
\usepackage[table,xcdraw]{xcolor}
\usepackage{caption}
\usepackage{float}
\usepackage{placeins}
\usepackage{color, colortbl}
\usepackage{stfloats}
\usepackage{enumitem}
\usepackage{tabularx}
\usepackage{xstring}
\usepackage{multirow}
\usepackage{xspace}
\usepackage{url}
\usepackage{subcaption}
\usepackage{xcolor}
\usepackage[hang,flushmargin]{footmisc}

\ifcamera \usepackage[accsupp]{axessibility} \fi

% ...

\ifarxiv  \fi

\newcommand{\R}[1]{{%
    \textbf{%
        \ifstrequal{#1}{1}{\textcolor{red}{R#1}}{%
        \ifstrequal{#1}{2}{\textcolor{blue}{R#1}}{%
        \ifstrequal{#1}{3}{\textcolor{magenta}{R#1}}{%
        \ifstrequal{#1}{4}{\textcolor{teal}{R#1}}{%
                           \textcolor{cyan}{R#1}%
        }}}}%
    }%
}}

\usepackage{xr-hyper}
\usepackage{multirow}
\usepackage{makecell}
\usepackage{cite}

\makeatletter
\newcommand*{\addFileDependency}[1]{
  \typeout{(#1)}
  \@addtofilelist{#1}
  \IfFileExists{#1}{}{\typeout{No file #1.}}
}

\makeatother

\usepackage[pagebackref,breaklinks,colorlinks]{hyperref}
\usepackage[capitalize]{cleveref}
\crefname{section}{Sec.}{Secs.}
\crefname{table}{Table}{Tables}
\crefname{figure}{Fig.}{Figs.}

\begin{document}

\definecolor{deemph}{gray}{0.6}
\newcommand{\gc}[1]{\textcolor{deemph}{#1}}
\newcommand{\cmark}{\ding{51}\xspace}%
\newcommand{\cmarkg}{\textcolor{lightgray}{\ding{51}}\xspace}%
\newcommand{\xmark}{\ding{55}\xspace}%
\newcommand{\xmarkg}{\textcolor{lightgray}{\ding{55}}\xspace}%

%% TITLE
\title{A Simple and Generic Framework for Feature Distillation via Channel-wise Transformation}
\author{\authorBlock}
\renewcommand{\thefootnote}{\fnsymbol{footnote}}
\setcounter{footnote}{1}
\maketitle

\begin{abstract}

Knowledge distillation is a popular technique for transferring the knowledge from a large teacher model to a smaller student model by mimicking. However, distillation by directly aligning the feature maps between teacher and student may enforce overly strict constraints on the student thus degrade the performance of the student model. To alleviate the above feature misalignment issue, existing works mainly focus on spatially aligning the feature maps of the teacher and the student, with pixel-wise transformation. In this paper, we newly find that aligning the feature maps between teacher and student along the channel-wise dimension is also effective for addressing the feature misalignment issue. Specifically, we propose a learnable nonlinear channel-wise transformation to align the features of the student and the teacher model. Based on it, we further propose a simple and generic framework for feature distillation, with only one hyper-parameter to balance the distillation loss and the task specific loss. Extensive experimental results show that our method achieves significant performance improvements in various computer vision tasks including image classification (+3.28\% top-1 accuracy for MobileNetV1 on ImageNet-1K), object detection (+3.9\% bbox mAP for ResNet50-based Faster-RCNN on MS COCO), instance segmentation (+2.8\% Mask mAP for ResNet50-based Mask-RCNN), and semantic segmentation (+4.66\% mIoU for ResNet18-based PSPNet in semantic segmentation on Cityscapes), which demonstrates the effectiveness and the versatility of the proposed method. The code will be made publicly available.

\end{abstract}

\section{Introduction}
    
    Nowadays, the development of deep neural network (DNN) architectures, such as ResNet~\cite{resnet}, ResNeXt~\cite{Xie2016resnext}, Faster R-CNN~\cite{ren2015faster}, and PSPNet~\cite{zhao2017pspnet}, has led to significant performance improvements for various computer vision tasks, such as image classification, object detection, and semantic segmentation.  
    However, the high performance of these DNN models comes at the cost of large size and high computational requirements for these architectures, which poses challenges for deployment of them in resource-constrained environments.
    \begin{table}[t]
\centering
\setlength{\tabcolsep}{4pt}
\begin{tabular}{lccc}
\toprule
\textbf{Task} & \textbf{Cls} & \textbf{Det} & \textbf{Seg} \\
\textbf{Metric} & \textbf{Top-1 Acc} & \textbf{BBox mAP} & \textbf{mIoU} \\
\midrule
\textbf{\gc{Student}} & \gc{69.9} & \gc{36.5} & \gc{69.9} \\
\textbf{\gc{Teacher}} & \gc{73.6} & \gc{41.0} & \gc{75.9} \\
\midrule
\textbf{Identity} & 70.3 (+0.4) & 38.8 (+2.3) & 46.2 (-23.7) \\
\textbf{Linear} & 71.0 (+1.1) & 39.3 (+2.8) & 71.4 (+1.5) \\
\textbf{Task-Specific$^{*}$} & 70.9 (+1.0) & 39.3 (+2.8) & 72.4 (+2.5) \\
\textbf{MLP(ours)} & 71.5 (+1.6) & 39.5 (+3.0) & 73.5 (+3.6) \\
\bottomrule
\multicolumn{4}{l}{$^{*}$ We use TaT~\cite{lin2022tat} in Cls and Seg, FGD~\cite{yang2022fgd} in Det}\\
\end{tabular}
\vspace{-2mm}
\caption{
% Comparison of various transformation methods in knowledge distillation across different computer vision tasks, including classification(Cls), Segmentation(Seg) and Detection(Det). 
Comparison of various transformation methods in knowledge distillation for classification(Cls), Segmentation(Seg) and Detection(Det) tasks. 
%The "Direct" approach directly mimics features using the l2-distance metric, the "Linear" approach employs a linear conv1x1 transformation,
%while the "Task-Specific" method utilizes established techniques specifically designed for each task. % such as TaT-Cls~\cite{lin2022tat} for classification, FGD~\cite{yang2022fgd} for object detection, and TaT-Seg~\cite{lin2022tat} for semantic segmentation.
% Further details can be found in the Appendix.
% The results demonstrate that with equal channel numbers, channel-wise transformation can still improves performance, and direct feature mimicking may reduce performance in some cases
Teacher and student feature maps have the same number of channels. 
Distillation with the help of the transformation module can improve student performance compared to direct mimics. 
}
\vspace{-2mm}
\label{table:DifferentTasks}
\end{table}
    To address this problem, knowledge distillation~\cite{hinton2015distilling} has been proposed to achieve high performance with reduced computational cost by transferring the knowledge from a large model (teacher) to a smaller model (student).
    
    Specifically, feature-based knowledge distillation methods, which transfer knowledge from the intermediate layer features of the teacher model to the student model, have been intensively studied and demonstrated as a more effective and generic approach for improving the performance of student model.
    
    As point out in~\cite{lin2022tat}, due to the feature misalignment of the teacher and student model, directly mimicking the intermediate features of the teacher model via vanilla $L_2$ distances may enforce overly strict constraints on the student, leading to sub-optimal performance.
    
    To alleviate this problem, existing works design novel distillation loss functions~\cite{heo2019overhaul,shu2021cwd} or feature transformation modules~\cite{zhang2020fkd,yang2022fgd,lin2022tat,reviewkd,kzagoruyko2016at} to indirectly mimic the teacher’s features. Specifically, the latter kind of approaches often focus on the feature transformations along the spatial dimension, such as, guiding the student's attention towards the key regions of the feature map~\cite{kzagoruyko2016at} or the relationship between different pixels~\cite{zhang2020fkd,yang2022fgd,lin2022tat}.
    
     In this paper, we focus on the feature-based knowledge distillation and make an effort to address the feature misalignment problem along the channel dimension rather than spatial dimensions. 
     We have observed that channel-wise transformations (e.g., 1x1 convolution) have been widely used to align the features of different channel sizes in many tasks including the feature-based knowledge distillation. Moreover, for the feature-based knowledge distillation task, these channel-wise transformation module are discarded when the channel sizes of the teacher’s feature and student’ s feature are already the same. However, we empirically find that a linear channel-wise transformation, \emph{i.e.}, 1x1 convolution , can result in consistent performance improvements for feature-based knowledge distillation, even when the channel sizes of teacher’s feature and student’ s feature are already the same, the results are shown in table~\ref{table:DifferentTasks}. 
     
     Inspired by our empirical findings about the importance of channel-wise transformations for feature-based distillation, we propose a simple and generic approach that focuses on channel-wise feature alignment. Specifically, without careful selection or design of transformation modules, we implement the channel-wise transformation as a non-linear MLP with one hidden layer, which has been demonstrated to have universal approximation capabilities~\cite{cybenko1989approximation}. Together with this simple channel-wise transformation module and the conventional $L2$-distance loss, we propose a very simple and generic method for feature-based distillation. 
     With only one tun-able hyper-parameter, our method is easy to apply to different tasks.
     
     Our extensive evaluation, as shown in Table~\ref{table:Comparewithsota}, reveals that our method consistently outperforms existing feature-based distillation methods on dense prediction tasks. In object detection, we observed consistent performance gains over two-stage, anchor-based, and anchor-free single-stage detectors, with an average improvement of +3.5\% in bbox mAP across these settings. For semantic segmentation, our method delivered an average improvement of +4.0\% in mIoU over heterogeneous and homogeneous distillation settings on the ResNet-18-based PSPNet. Our method also achieves strong performance on the classification task, with an average increase of +2.4\% in Top-1 accuracy, regardless of whether the number of channels in the student and teacher feature maps are the same or not. 
     \begin{table}[t]
\centering
% \small
\setlength{\tabcolsep}{5pt}
\begin{tabular}{ccccccc}
\toprule
& & Cls & Det & Ins Seg & Seg & \#Hyper \\
\midrule
KR\cite{reviewkd} & & \bf{+2.5} & - & - & - & 2\\
FGD\cite{yang2022fgd} & & - & +3.1 & +2.4 & - & 5 \\
CWD\cite{shu2021cwd} & & - & - & - & +3.2 & 2\\
MGD\cite{yang2022mgd} & & +2.4 & +3.3 & +2.7 & +3.3 & 2\\ \midrule
\textbf{Ours} & & +2.4 & \bf{+3.5} & \bf{+2.8} & \bf{+4.0} & \bf{1}\\
\bottomrule
\end{tabular}
\caption{
% A comparison of our method with other state-of-the-art methods on various computer vision tasks, including image classification, object detection, instance segmentation, and semantic segmentation. The number of hyperparameters is shown in the column "\#Hyper". The metrics reported are the average improvement in Top-1 accuracy (Cls), BBox mAP (Det), Mask AP (Instance Seg), and mIoU (Seg). Our method achieves state-of-the-art results while being simple and efficient, requiring fewer hyperparameters.
Comparisons of the state-of-the-art methods on image classification (Cls), object detection (Det), instance segmentation (Ins Seg), and semantic segmentation (Seg). The metrics reported are Top-1 accuracy, BBox mAP, Mask AP, and mIoU, improvement relative to students, respectively. Hyper denotes hyperparameters. 
Our method achieves state-of-the-art results with only 1 hyperparameter.
% Our method achieves state-of-the-art results while being simple and efficient, requiring fewer hyperparameters.
}
\label{table:Comparewithsota}
\vspace{-2mm}
\end{table}

To sum up, our main contributions are three-folds:

\begin{itemize}
    
    \item We reinstate the importance of channel-wise transformation for aligning the student's and teacher's features in feature-based knowledge distillation.
    \item We propose a simple and generic framework for feature-based knowledge distillation which uses MLP as the channel-wise transformation module to help student learn more powerful features.
    \item We achieve state-of-the-art distillation results for multiple dense prediction tasks, and comparable state-of-the-art results for classification task.
\end{itemize}

\label{sec:intro}

\section{Related Work}
    The concept of knowledge distillation was first proposed by Hinton et al. ~\cite{hinton2015distilling}, with the goal of transferring dark knowledge from a cumbersome teacher model to a smaller student model to improve the student's performance. Based on the types of dark knowledge, mainstream knowledge distillation methods can be divided into two categories: Logits-based knowledge distillation and feature-based knowledge distillation.
\subsection{Logits-based knowledge distillation}
    Classical logits-based knowledge distillation methods~\cite{hinton2015distilling} minimize the KL divergence between the output logits of teacher and student models. One recent line of research focuses on refining the vanilla knowledge distillation loss function to better leverage the logits information. WSLD~\cite{zhou2020wsld} rethinks the knowledge distillation process from a bias-variance trade-off perspective and proposes weighted soft labels for knowledge distillation. DKD~\cite{zhao2022dkd}, reformulates the classical knowledge distillation loss into the target and non-target part and calculates the distillation loss separately. While these works have improved the performance of logits-based knowledge distillation methods on classification tasks, they have often not achieved significant results on other tasks, such as dense prediction tasks.
%Head-related:   
    %However, this distillation manner is related to the classification head, for more sophisticated tasks such as object detection which involves the classification task and the regression task, Logits-based methods often fail to achieve ideal performance due to the task-specific property such as the imbalanced distribution of pixels between fore-groud and background.
%Trial on different tasks: 
%Recently, there are some works that try to use the logit-based knowledge distillation methods on other tasks, which often consider a reformulation of traditional Knowledge distillation loss.
    Another line of work involves modeling other tasks into a classification task, and adopting the logits-based knowledge distillation on other tasks.
        LD\cite{zheng2022ld}, reformulates the output form of the regression head to a probability distribution and applies classical knowledge distillation to the regression task. However, it is only for object detection tasks and requires changes to the detection head. RMKD\cite{li2022rm} reformulates the ordering between anchors into the form of the probability distribution for knowledge transfer and applies classical knowledge distillation to the regression task. However, it is only limited to anchor-based detectors. 
\subsection{Feature-based knowledge distillation}
% \paragraph{Feature-based knowledge distillation for classification}

\paragraph{For classification.}
The feature of the teacher model is another kind of dark knowledge and was first used in~\cite{romero2014fitnets}. Subsequent works have primarily focused on finding more effective ways to utilize this type of dark knowledge.
    AT\cite{kzagoruyko2016at} extracts attention maps to help the student to pay attention to the import regions. However, it squeezes the channel dimension of the feature map and fails to utilize the channel information, resulting in limited improvements to the student models
    OFD\cite{heo2019overhaul} designs a new loss function and used marginal ReLU to extract the major information in the network. 
    CRD\cite{tian2019crd} incorporates the idea of contrastive learning into knowledge distillation, and although it has achieved relatively good performance, However, its training cost is high due to the use of memory banks for a large amount of negative samples.
    KR\cite{reviewkd} proposes conducting knowledge distillation on multi-level features in a review manner,resulting in good performance, especially on classification tasks.
    TaT~\cite{lin2022tat} propose a novel one-to-all spatial matching approach for knowledge distillation based on similarity generated from a target-aware transformer.
% \paragraph{Feature-based knowledge distillation for dense prediction}

\paragraph{For objection detection.}
% For objection task:
    Object detection is a significantly more complex task than image classification. The extreme imbalance between foreground and background pixels poses a major challenge for object detection. To address this issue, many knowledge distillation methods attempt to have the student model imitate the key regions of the teacher model.
    FGFI\cite{wang2019fgfi} leverages fine-grained masks to force students to focus on foreground regions.
    GID\cite{dai2021GID} identifies regions where the student and teacher models perform differently as the key regions for distillation, without relying on anchors.
    Defeat\cite{guo2021defeat} finds that the background region also contains valuable information and proposes distilling foreground and background regions separately.
    Recent methods have also discovered that the relationships between different pixels are important knowledge for distillation and propose various global modules to address this problem. 
    FKD\cite{zhang2020fkd} employs attention masks to direct the student model's focus on key regions and non-local modules to capture the relationships between different pixels, resulting in improved knowledge distillation.
    FGD\cite{yang2022fgd} proposes focal and global distillation mechanisms, forcing the student model to learn the teacher's crucial region and global information through a global context block~\cite{cao2019gcnet}

\vspace{-4mm}
\paragraph{For semantic segmentation.}

\begin{figure*}
    \centering
     \begin{subfigure}[b]{0.45\textwidth}
         \centering
         \includegraphics[width=0.95\textwidth]{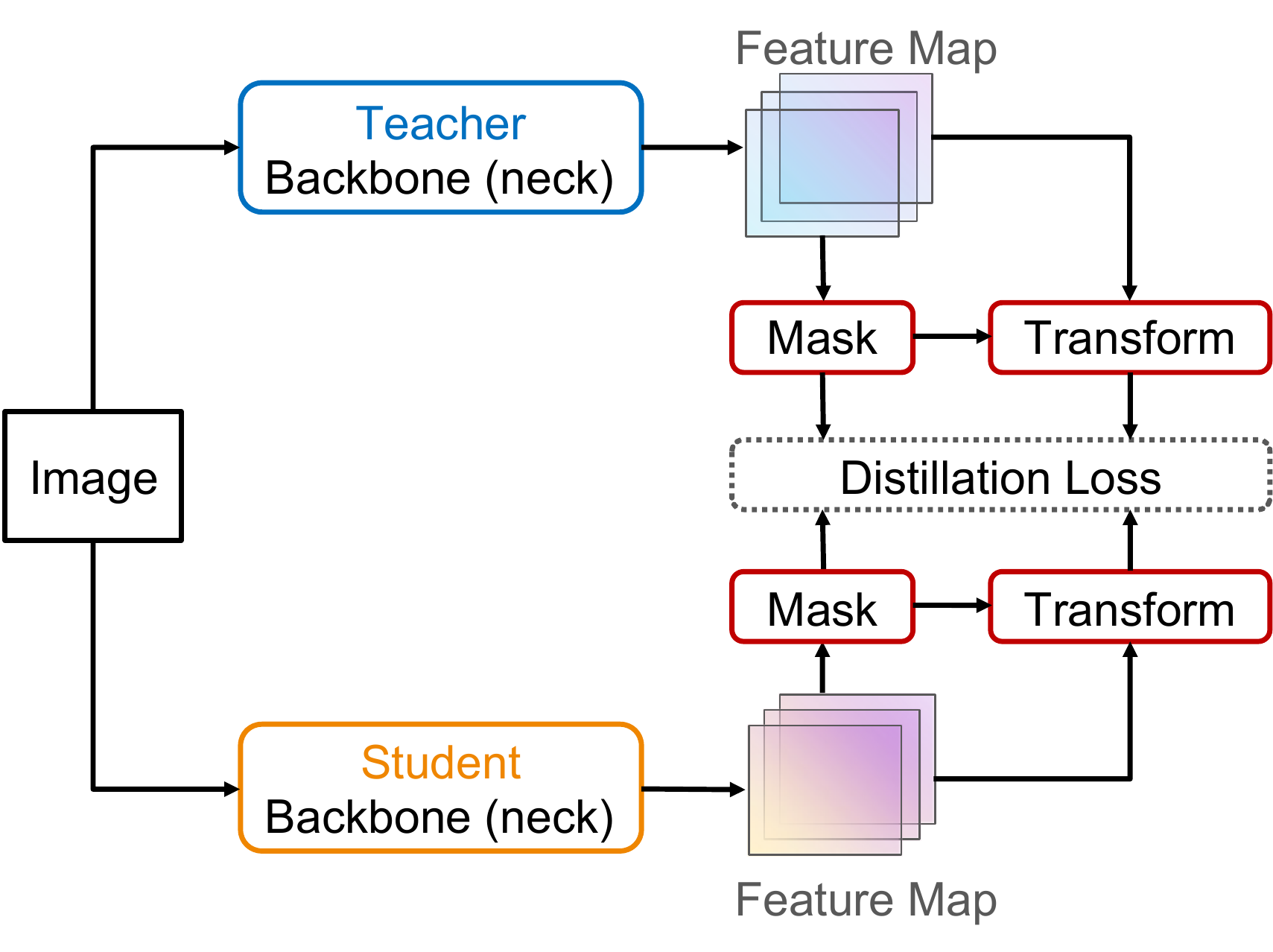}
         \caption{Some previous methods~\cite{zhang2020fkd,yang2022fgd} use both sophisticated designed mask and spatial-wise transformation for both teacher and student.}
         \label{fig:fkd}
     \end{subfigure}
     \hfill
     \begin{subfigure}[b]{0.48\textwidth}
         \centering
         \includegraphics[width=0.95\textwidth]{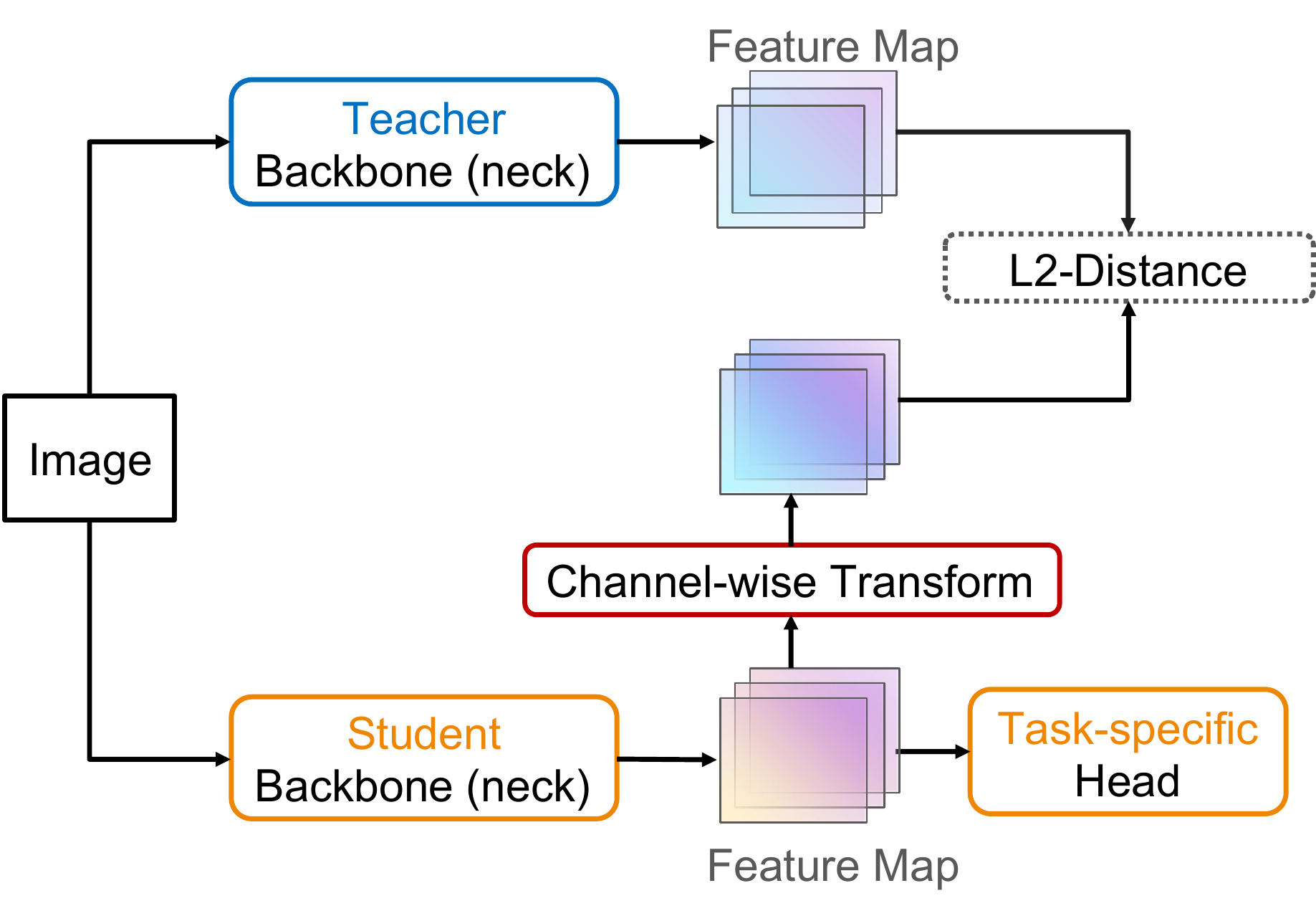}
         \caption{Our proposed method uses learn-able channel-wise transformation only for the student model.}
         \label{fig:sgkd}
     \end{subfigure}
     \vspace{-2mm}
    \caption{Difference between our method and existing feature-based knowledge distillation methods.}
    % \vspace{-2mm}
    
    \label{fig:my_label}
\end{figure*}

% For semantic segmentation:
    Semantic segmentation is a per-pixel prediction problem, and strictly aligning the feature maps between the student and teacher models may impose overly strict constraints and lead to sub-optimal results.~\cite{shu2021cwd}. Recent works~\cite{liu2019skds,wang2020ifvd} try to force the student to learn the correlations among different spatial regions.
    IFVD\cite{wang2020ifvd} focuses on the intra-class feature variation among pixels with the same label and designs an IFV module to transfer the structural knowledge. 
    SKDS~\cite{liu2019skds} combines pixel-wise distillation, pair-wise distillation, and holistic distillation using a GAN-based approach to align the output maps of teacher and student models.
    CIRKD\cite{yang2022cirkd} aims to model the pixel-to-pixel and pixel-to-region relationships as supervisory signals for knowledge distillation in the semantic segmentation task.
    CWD~\cite{shu2021cwd} derives probability maps by normalizing the activation maps of each channel of intermediate features, and minimizes the KL divergence between these probability maps, applying the method to dense prediction tasks, including object detection and semantic segmentation. 
\vspace{-4mm}
\paragraph{For general tasks.}
% General:
MGD~\cite{yang2022mgd} employs a generative approach that involves the use of  random masks that randomly to erases a portion of the student's feature map and then forces it to generate features similar to the teacher's through an adversarial generator and applies it to classification, detection, and segmentation tasks.  

In this paper, we focus on channel-wise transformations, and propose a simple and generic method for feature-based knowledge distillation. 
\label{sec:related}

\section{Method}

In this section, we first briefly introduce the basic form of intermediate feature-based knowledge distillation, and then present the details of our proposed method.

\subsection{Revisiting Feature-based Knowledge Distillation}
In feature-based knowledge distillation, a student model is generally supervised by a teacher model as~\cite{gou2021knowledge} :
% Generally, the pipeline for feature-based knowledge distillation can be formulated as:
\begin{equation}
L_{feat} = \mathcal{L}_{KD}\left(\mathcal{T}_t\left(\boldsymbol{F}_t\right), \mathcal{T}_s\left(\boldsymbol{F}_s\right)\right),
\end{equation}\label{egu:kd}
% \begin{equation}
% \mathcal{L}_{\text {mimic }}=\frac{1}{H W C}\left\|\boldsymbol{F}^{(t)}-\phi\left(\boldsymbol{F}^{(s)}\right)\right\|_2^2
% \end{equation}
where  $\mathcal{L}_{KD}$ represents the similarity function used to match the feature maps of the teacher model, $\bf{F}_t$, and the student model, $\bf{F}_s$. Furthermore, the transformation functions, $\mathcal{T}_t$ and $\mathcal{T}_s$, will be performed when the feature maps of the teacher and student models are not of the same shape (e.g., a linear projection layer to align the number of channels in $\bf{F}_s$ with those in $\bf{F}_t$).

%Early works in knowledge distillation introduced various types of similarity function $\mathcal{L}_{KD}$, such as Kullback–Leibler (KL) divergence~\cite{shu2021cwd} and maximum mean discrepancy~\cite{huang2017like}. For example, CWD~\cite{shu2021cwd} uses KL divergence between the channel-wise probability maps of the student and teacher models as the similarity function $\mathcal{L}_{KD}$. On the other hand,
%recent works introduce complex transformation ($\mathcal{T}_t$ and $\mathcal{T}_s$) to guide and help students for learning knowledge (aligning feature) from teachers. For example, as shown in Figure~\ref{fig:fkd}, FKD~\cite{zhang2020fkd} and FGD~\cite{yang2022fgd} exploit (1) specific module (i.e., Non-local module~\cite{wang2018non} or GCBlock~\cite{cao2019gcnet}) and (2) channel and spatial attention masks applied to both the teacher and student features. 
%Recent works utilize complex transformations ($\mathcal{T}_t$ and $\mathcal{T}_s$) to assist students in learning knowledge (feature alignment) from teachers. For instance, as depicted in Figure~\ref{fig:fkd}, FKD~\cite{zhang2020fkd} and FGD~\cite{yang2022fgd} employ (1) specific modules (\textit{i.e}., Non-local module~\cite{wang2018non} or GCBlock~\cite{cao2019gcnet}) and (2) channel-wise and spatial-wise attention masks to the teacher and student features.
%This raises a question: 
%Do student networks really need the assistance of well-designed modules to learn better features from the teacher?
Recently, several works have employed complex transformations ($\mathcal{T}_t$ and $\mathcal{T}_s$) to facilitate the acquisition of knowledge (feature alignment) by student networks from teacher networks. For example, as depicted in Figure~\ref{fig:fkd}, both FKD~\cite{zhang2020fkd} and FGD~\cite{yang2022fgd} utilize (1) specific modules, such as the Non-local module~\cite{wang2018non} or GCBlock~\cite{cao2019gcnet}, and (2) channel-wise and spatial-wise attention masks to align the features of the teacher and student networks.
This raises the question of whether the use of well-designed modules is necessary for student networks to learn more effective features from the teacher.

\begin{figure*}
    \centering
     \begin{subfigure}[b]{0.35\textwidth}
         \centering
         \includegraphics[width=0.67\textwidth]{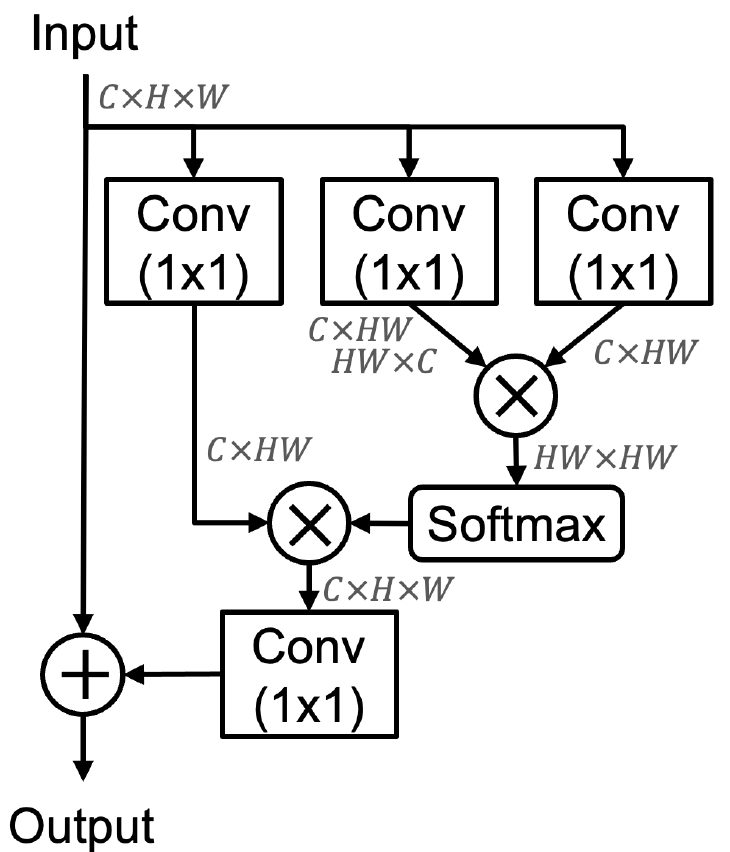}
         \caption{Non-Local Block used in FKD~\cite{zhang2020fkd}}
         \label{fig:transform_nl}
     \end{subfigure}
     \hfill
     \begin{subfigure}[b]{0.35\textwidth}
         \centering
         \includegraphics[width=0.7\textwidth]{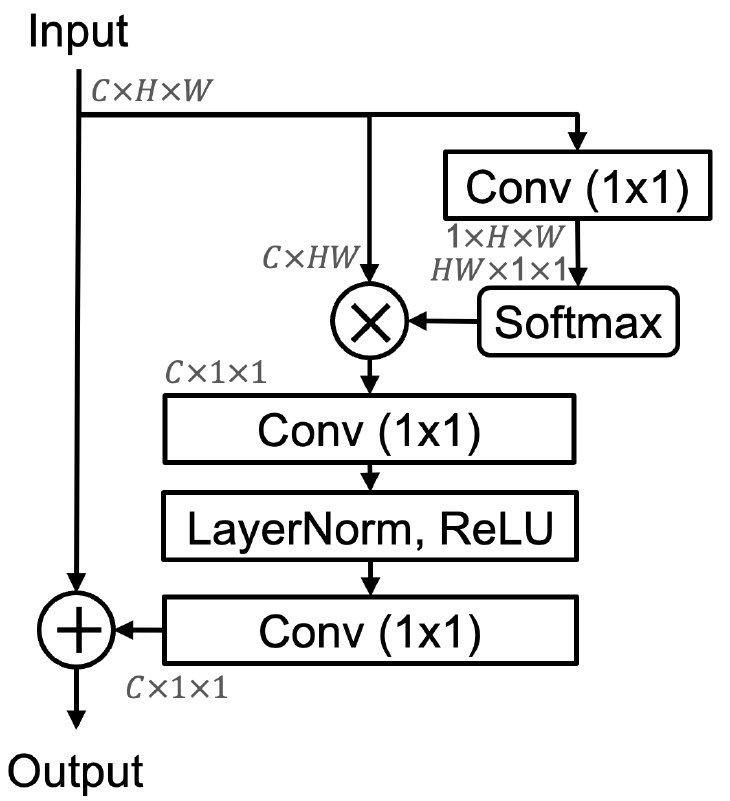}
         \caption{Global Context Block used in FGD~\cite{yang2022fgd}}
         \label{fig:transform_gc}
     \end{subfigure}
     \hfill
     \begin{subfigure}[b]{0.2\textwidth}
         \centering
         \includegraphics[width=0.57\textwidth]{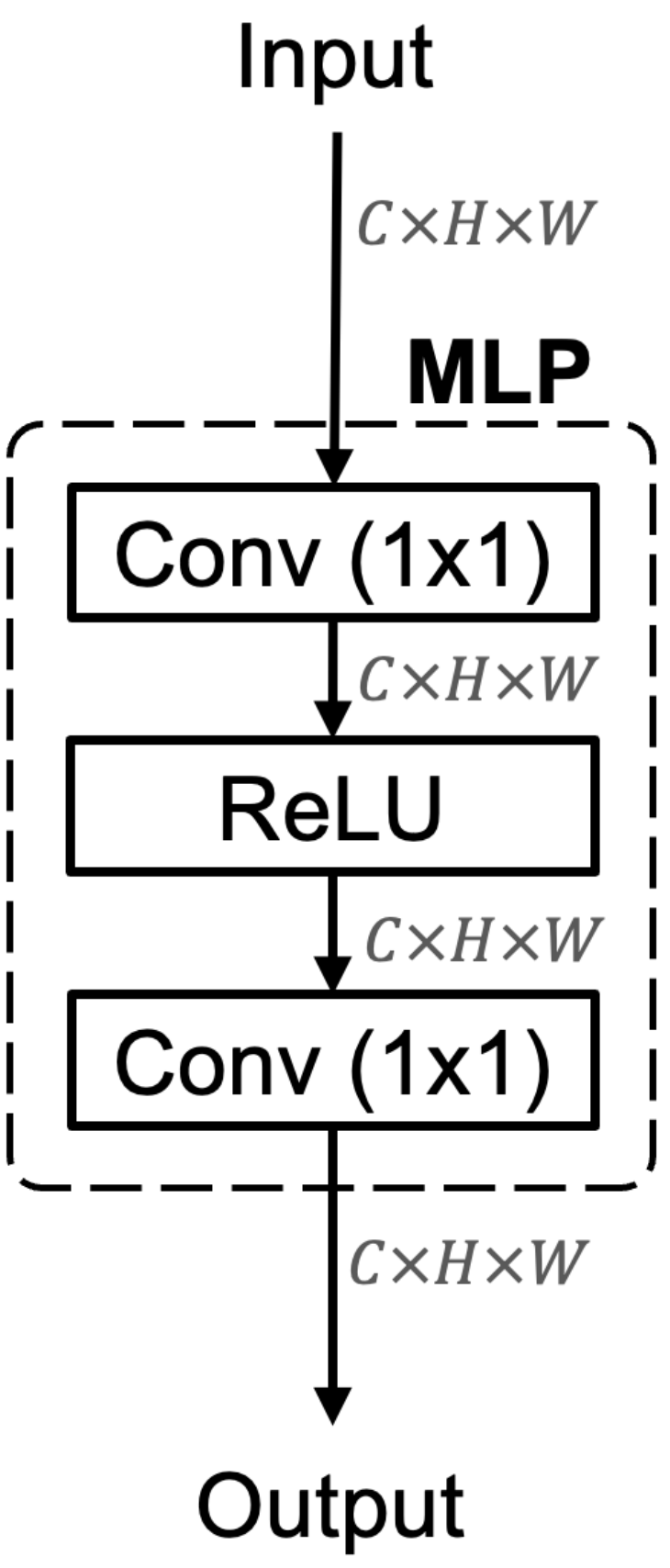}
         \caption{MLP (Ours)}
         \label{fig:transform_mlp}
     \end{subfigure}
    \caption{Comparison of different transformation modules in knowledge distillation. FKD~\cite{zhang2020fkd} uses a non-local module (a), while FGD~\cite{yang2022fgd} employs GCBlock (b) to model the relationships between pixels in an image. Our method utilizes a simple yet effective channel-wise transformation through an MLP (c), consisting of two 1$\times$1 convolution layers and a ReLU activation layer.}
    \vspace{-2mm}
    \label{fig:transform}
\end{figure*}

In this paper, we perform empirical studies to address the question raised above and find that student models can enhance their representations through a straightforward nonlinear channel-wise transformation. Based on this finding, as illustrated in Figure~\ref{fig:sgkd}, we introduce a simple method that incorporates a Multi-Layer Perceptron (MLP) into the student features and aligns the transformed student features with the teacher features using a conventional $L_2$ distance. The specifics of our proposed method are outlined in the subsequent subsection.

\subsection{Learnable channel-wise transformation}
Instead of using complex transformation on both spatial and channel dimension, we propose to use a learnable nonlinear channel-wise transformation to align the feature maps of the student and the teacher model. In detail, we use a non-linear MLP with one hidden layer for the student (\textit{i.e.}, $\mathcal{T}_t = identity$ and $\mathcal{T}_s = MLP$ in Equ.~\ref{egu:kd}):

\begin{equation}
\operatorname{MLP}(F)=W_{2}\left(\sigma \left(W_{1}(F)\right)\right),
\end{equation}

where $W1$ and $W2$ are learnable parameters implemented as 1$\times$1 convolutions, and $\sigma$ represents ReLU activation. As illustrated in Figure~\ref{fig:transform}, our transformation module (Figure~\ref{fig:transform_mlp}) is much simpler than the methods proposed in  FKD~\cite{zhang2020fkd} and  FGD~\cite{yang2022fgd}.

Without bells and whistles, we choose $L_2$ distance for supervising transformed student feature and teacher feature. Specifically, the feature distillation loss is formulated as:
     \begin{equation}
        L_{feat} = \sum_{i}^{N}(MLP(\boldsymbol{F}_s ) - \boldsymbol{F}_t)^2.
    \end{equation}

\begin{algorithm}[t]
\caption{Pseudo code of our method in a PyTorch-like style.}
\label{alg:code}
% \algcomment{\fontsize{9pt}{0em}\selectfont \texttt{bmm}: batch matrix multiplication; \texttt{mm}: matrix multiplication; \texttt{cat}: concatenation.
% %\vspace{-1.em}
% }
\definecolor{codeblue}{rgb}{0.25,0.5,0.5}
\lstset{
  backgroundcolor=\color{white},
  basicstyle=\fontsize{9pt}{9pt}\ttfamily\selectfont,
  columns=fullflexible,
  breaklines=true,
  captionpos=b,
  commentstyle=\fontsize{9pt}{9pt}\color{codeblue},
  keywordstyle=\fontsize{9pt}{9pt},
%  frame=tb,
}
\begin{lstlisting}[language=python]
# f_mlp: 2-layer MLP with ReLU activation
# x_s: student feature [N, C, H, W]
# x_t: teacher feature [N, C, H, W]

def forward(x_s, x_t):
    n = x_s.shape[0] # size of mini-batch
    
    # channel-wise non-linear transformation
    x_mlp = f_mlp.forward(x_s) 
    
    # calculate l2 distance
    diff = (x_mlp - x_t).pow(2) 
    
    # distillation loss (averaged by batch)
    loss = diff.sum() / n
    
    return loss
\end{lstlisting}
\end{algorithm}    

Our approach is straightforward and can be efficiently executed using prevalent machine learning libraries, such as PyTorch, as shown in Algorithm~\ref{alg:code}. The ease of implementation enables us to leverage existing infrastructure, facilitating the training process of our model.

\subsection{Overall loss}
Our method can be easily used in various tasks. Combined with task-specific losses, the overall loss can be formulated as:
\begin{equation}
     L_{total} = L_{task} + \alpha L_{feat}
\end{equation}
where $\alpha$ is a hyper-parameter to balance the weight of knowledge distillation loss.

\label{sec:method}

\section{Experiment}
Our approach, which is a feature-based method, is easy to implement on various models and tasks. In this paper, we conduct experiment on image classification, object detection, instance segmentation, and semantic segmentation, to demonstrate the simplicity, effectiveness, and generality of our method.
\subsection{Image Classification on ImageNet}
\begin{table*}

    \centering
  \begin{tabular}{l|l|c|c|l|c|c}
    \toprule
 Mechanism & Method & \makecell{Top-1 \\ acc}  & \makecell{Top-5 \\ acc} & Method & \makecell{Top-1 \\ acc} & \makecell{Top-5 \\ acc} \\
    \midrule
- & ResNet-50(T) &76.55 & 93.06 & ResNet-34(T) & 73.62 &91.59  \\ 
- & MobileNet(S) & 69.21 & 89.02 & ResNet-18(S) & 69.90 &89.43  \\
\midrule
     \multirow{2}{*}{\makecell{Logits}} &KD\cite{hinton2015distilling} & 70.68 & 90.30 & KD\cite{hinton2015distilling} & 70.68 & 90.16 \\
    &DKD\cite{zhao2022dkd} & 72.05 & 91.0 & DKD$^{*}$\cite{zhao2022dkd} & 71.37 & 90.26  \\
    \midrule
    \multirow{3}{*}{\makecell{Multi Feature}} 
    &AT\cite{kzagoruyko2016at} & 70.72 & 90.03 & AT\cite{kzagoruyko2016at} & 70.59 & 89.73  \\
    &OFD\cite{heo2019overhaul} & 71.25 & 90.34 & OFD\cite{heo2019overhaul} & 71.08 & 90.07 \\
    &KR\cite{reviewkd} & 72.56 & 91.00 & KR\cite{reviewkd} & 71.61 & 90.51 \\
    \midrule
    \multirow{4}{*}{\makecell{Single Feature}} 
    &RKD\cite{rkd} & 71.32 & 90.62 & RKD\cite{rkd} & 71.34 & 90.37 \\
    &CRD\cite{crd} & 71.40 & 90.42 & CRD\cite{crd} & 71.17 & 90.13 \\
    & MGD\cite{yang2022mgd} & 72.35 & 90.71 & MGD\cite{yang2022mgd} & 71.58 & 90.35 \\
    &\cellcolor{lightgray!45}Ours & \cellcolor{lightgray!45}72.49&\cellcolor{lightgray!45}90.81 &\cellcolor{lightgray!45}Ours & \cellcolor{lightgray!45}71.51&\cellcolor{lightgray!45}90.32 \\
    \bottomrule
    % \multicolumn{7}{c}{$^{*}$ We report the result inplemented in MMRazor\cite{2021mmrazor}.} \\
  \end{tabular}
  \caption{Results of different distillation methods on ImageNet dataset for the image classification task. {\bf T} and {\bf S} mean the teacher and student, respectively. $^{*}$ We report the result inplemented in MMRazor\cite{2021mmrazor}.}
  \vspace{-0.05in}
  \label{table:classification results}
\end{table*}

\paragraph{Settings}
	To evaluate our method on the image classification task, we use the Imagenet\cite{deng2009imagenet} dataset. Specifically, we use 1.2 million images from the Imagenet training set to train the model and 50,000 images from the validation set to test the model and use Top-1 accuracy as the evaluation metric. We use a standard training procedure, which involves training the model for 100 epochs, with a learning rate decay at the 30th, 60th, and 90th epochs, The optimizer used for training is SGD, and the initial learning rate is set to 0.1. The training is performed on 8 GPUs with a batch size of 32 images per GPU.
% 	We select two sets of models setting a is resnet50 as the teacher model and MobileNet as the student model, setting b is resnet34 as the teacher model and resnet18 as the student model, both setting a and setting b is on the output of the feature from the last stage of model backbone Calculate distillation loss, distillation loss weight $\alpha$ is set to $7\times10^{-5}$.

    We evaluate our method on both homogeneous and heterogeneous distillation settings. Specifically, we test our method on two model configurations: a) ResNet34 as the teacher model and ResNet18 as the student model, and b) ResNet50 as the teacher model and MobileNet as the student model. In both configurations, we use the feature maps from the last stage of the backbone to calculate the distillation loss, and set the distillation loss weight $\alpha$ to $7\times10^{-5}$. Moreover, our method is compared with not only single layer feature-based methods\cite{park2019rkd,tian2019crd,yang2022mgd} but also other state-of-the-art logits-based methods\cite{hinton2015distilling,zhao2022dkd} and multi-layer feature-based methods\cite{kzagoruyko2016at,heo2019overhaul,reviewkd}.
% 	We compare our method not only to single-feature-based methods but also to other state-of-the-art methods. 
% 	For a fair comparison, we only compare the methods that train under the same settings.
	
\paragraph{Comparison to baseline.} 
As presented in Table~\ref{table:classification results}, our method demonstrates its effectiveness on the image classification task. Specifically, under the homogeneous setting, the Top-1 accuracy of ResNet-18 is improved by +3.28\%. Similarly, under the heterogeneous setting, the Top-1 accuracy of MobileNet is improved by +1.61\%. These results highlight the superiority of our method in comparison to the baseline models.

\paragraph{Comparison to single feature distillation methods}
    Our method outperforms all single-feature distillation methods\cite{park2019rkd,tian2019crd,yang2022mgd} in the heterogeneous setting and is on par with the state-of-the-art method MGD\cite{yang2022mgd}. These results highlight the effectiveness of our method in comparison to other single-feature distillation techniques.
\paragraph{Comparison to previous state-of-the-art}
    Compared to the previous state-of-the-art method KR~\cite{reviewkd}, which uses multi-stage features, our simple method achieves comparable performance with a difference of less than 0.1\% in terms of Top-1 accuracy on both homogeneous and heterogeneous settings.
\label{sec:cls_experiment}
% \subsection{Object Detection and Instance Segmentation}
\subsection{Object Detection on COCO}
\paragraph{Settings} For the object detection task, we evaluate our method on the COCO\cite{lin2014coco} dataset. Specifically, we use 120,000 images from the COCO training set for model training and 5,000 images from the validation set for model testing, with mAP as the evaluation metric. Our training procedure follows a standard 2x schedule, consisting of 24 training epochs, with the reduction of the learning rate at epochs 16 and 22. The optimization process is performed using Stochastic Gradient Descent (SGD) and the model is trained on 8 GPUs, each with a batch size of 2.

We experiment on multiple detector architectures, including two-stage, single-stage anchor-based, and single-stage anchor-free detectors. The distillation loss is computed on all feature maps output from the neck, and the distillation loss weight $\alpha$ is set to $5\times10^{-7}$ for the two-stage detector and $2\times10^{-5}$ for the one-stage detector. For the instance segmentation task, we use a ResNext-101-based Cascade Mask R-CNN as the teacher model and a ResNet-50-based Mask R-CNN as the student model. The experimental configuration follows that of the two-stage detector distillation.
\begin{table}[t]
    \centering
  \begin{tabular}{l|lc}
    \toprule
     Method & Input Size  & mIoU \\
    \midrule
    PspNet-Res101(T)&$512\times 1024$&78.34 \\
    PspNet-Res18(S) & $512\times 512$ &69.85 \\
    \midrule
    SKDS\cite{liu2019skds}  &$512\times 512$ & 72.70\\
    CWD\cite{shu2021cwd}  &$512\times 512$ & 73.53\\
    MGD\cite{yang2022mgd} &$512\times 512$ & 73.63\\
    \cellcolor{lightgray!45}Ours & \cellcolor{lightgray!45}$512\times 512$&\cellcolor{lightgray!45}74.51\\
    \midrule\midrule
    PspNet-Res101(T)&$512\times 1024$&78.34\\
    DeepLabV3-Res18(S) & $512\times 512$ &73.20\\
    \midrule
    SKDS\cite{liu2019skds} &$512\times 512$&73.87\\
    CWD\cite{shu2021cwd} &$512\times 512$&75.93\\
    MGD\cite{yang2022mgd} &$512\times 512$ & 76.02\\
    %MGD$^{*}$ &$512\times 512$ & 76.31\\
    \cellcolor{lightgray!45}Ours&\cellcolor{lightgray!45}$512\times 512$&\cellcolor{lightgray!45}76.55\\
    %\cellcolor{lightgray!45}Ours$^{*}$ & \cellcolor{lightgray!45}$512\times 512$&\cellcolor{lightgray!45}76.05\\
    \bottomrule
  \end{tabular}
  % \vspace{-1mm}
     \caption{Results of the semantic segmentation task on CityScapes dataset. {\bf T} and {\bf S} mean teacher and student, respectively.}
    
  \label{table:segmentation results}
\end{table}
\paragraph{Object Detection}
% \begin{table}[t]
%     \centering
%   \begin{tabular}{c|l|lccc}
%     \toprule
%     Teacher& Student & mAP \\
%     \midrule
%     \multirow{7}{*}{\makecell{RetinaNet\\ResNeXt101\\(41.0)}}
%     &RetinaNet-ResNet50 & 37.4 \\
%     &FKD\cite{zhang2020fkd} & 39.6 \\
%     &CWD\cite{shu2021cwd} &40.8 \\
%     &FGD\cite{yang2022fgd} &40.4 \\
%     &FGD$^{*}$ &40.7 \\
%     &MGD\cite{yang2022mgd}$^{*}$ &41.0 \\
%     &\cellcolor{lightgray!45}Ours & \cellcolor{lightgray!45}{41.0} \\
%     \midrule
%     \multirow{6}{*}{\makecell{Cascade\\Mask RCNN\\ResNeXt101\\(47.3)}}
%     &Faster RCNN-ResNet50 & 38.4 \\
%     &FKD  & 41.5 \\
%     &CWD &41.7 \\
%     &FGD\cite{yang2022fgd}&42.0 \\
%     &MGD &42.1 \\
%     &\cellcolor{lightgray!45}Ours & \cellcolor{lightgray!45}{42.3}\\
%     \midrule
%     \multirow{7}{*}{\makecell{Reppoints \\ResNeXt101\\(44.2)}}
%     &Reppoints-ResNet50 & 38.6 \\
%     &FKD  & 40.6 \\
%     &CWD &42.0 \\
%     &FGD\cite{yang2022fgd}&41.3 \\
%     &FGD$^{*}$ & 42.0 \\
%     &MGD$^{*}$ & 42.3\\
%     &\cellcolor{lightgray!45}Ours & \cellcolor{lightgray!45}{42.0}\\
    
%     \bottomrule
%   \end{tabular}
%   \caption{Results of different distillation methods for object detection on COCO.}
%   \vspace{-0.05in}
%   \label{table:detection results}
% \end{table}

\begin{table*}[t]
  \centering
  \begin{tabular}{c|l|cccc|cccc}
    \toprule
    Teacher& Student & mAP  & AP$_{S}$ & AP$_{M}$ &AP$_{L}$&mAR& AR$_{S}$ & AR$_{M}$ &AR$_{L}$\\
    \midrule
    \multirow{5}{*}{\makecell{RetinaNet\\ResNeXt101}}
    &RetinaNet-ResNet50 & 37.4 &20.6&40.7&49.7&53.9&33.1&57.7&70.2\\
    &FKD\cite{zhang2020fkd} & 39.6(+2.2)&22.7&43.3&52.5&56.1(+2.2)&36.8&60.0&72.1\\
    &FGD\cite{yang2022fgd}& 40.4(+3.0)&23.4&44.7&54.1&56.7(+2.8)&37.6&61.5&72.4\\
    &MGD\cite{yang2022mgd} & 40.6(+3.2) & 23.4 & 45.1 & 54.0 & 56.7(+2.8) & 37.1 & 61.0 & 72.5 \\
    &\cellcolor{lightgray!45}Ours & \cellcolor{lightgray!45}{41.0(+3.6)} & \cellcolor{lightgray!45}23.1 & \cellcolor{lightgray!45}45.5 & \cellcolor{lightgray!45}55.0 & \cellcolor{lightgray!45}56.8(+2.9) & \cellcolor{lightgray!45}37.2 & \cellcolor{lightgray!45}60.8 & \cellcolor{lightgray!45}72.4\\
    % \cline{2-10}
    % &\gc{CWD\dag \cite{shu2021cwd}} & \gc{40.8(+3.4)} &\gc{22.7} &\gc{44.5} &\gc{55.3} & \gc{-}&\gc{-}&\gc{-}&\gc{-} \\
    % &\gc{FGD\dag\cite{yang2022fgd}} & \gc{40.7(+3.3)}&\gc{22.9}&\gc{45.0}&\gc{54.7}&\gc{56.8(+2.9)}&\gc{36.5}&\gc{61.4}&\gc{72.8}\\
    %  &\gc{MGD\dag \cite{yang2022mgd}} &\gc{41.0(+3.6)} &\gc{23.4} &\gc{45.3}& \gc{55.7} & \gc{57.0(+3.1)} & \gc{37.2} & \gc{61.7} & \gc{72.8}\\
    \midrule
    \multirow{5}{*}{\makecell{Cascade\\Mask RCNN\\ResNeXt101}}
    &Faster RCNN-ResNet50 & 38.4 &21.5&42.1&50.3&52.0&32.6&55.8&66.1\\
    &FKD\cite{zhang2020fkd} & 41.5(+3.1)&23.5&45.0&55.3&54.4(+2.4)&34.0&58.2&69.9\\
    % &CWD\cite{shu2021cwd} & 41.7(+3.3) &23.3 &45.5 &55.5 & -&-&-&- \\
    &FGD\cite{yang2022fgd} &42.0(+3.6)&23.8&46.4&55.5&55.4(+3.4)&35.5&60.0&70.0\\
    &MGD\cite{yang2022mgd} &42.1(+3.7) &23.7 &46.4 &56.1 & 55.5(+3.5) & 35.4 & 60.0 & 70.5\\
    &\cellcolor{lightgray!45}Ours & \cellcolor{lightgray!45}{42.3(+3.9)} & \cellcolor{lightgray!45}24.2 & \cellcolor{lightgray!45}46.4 & \cellcolor{lightgray!45}56.1 & \cellcolor{lightgray!45}55.3(+3.3) & \cellcolor{lightgray!45}34.9 & \cellcolor{lightgray!45}59.8 & \cellcolor{lightgray!45}70.4\\
    \midrule
    \multirow{5}{*}{\makecell{RepPoints\\ResNeXt101}}
    &RepPoints-ResNet50 & 38.6&22.5&42.2&50.4&55.1&34.9&59.4&70.3\\
    &FKD\cite{zhang2020fkd} & 40.6(+2.0)&23.4&44.6&53.0&56.9(+1.8)&37.3&60.9&71.4\\
    &FGD\cite{yang2022fgd}& 41.3(+2.7)&24.5&45.2&54.0&58.4(+3.3)&39.1&62.9&74.2\\
    &MGD\cite{yang2022mgd}&
    41.7(+3.1)&24.1&45.8&55.3&57.9(+2.8)&39.0&62.0&73.6\\
    &\cellcolor{lightgray!45}Ours & \cellcolor{lightgray!45}{42.0(+3.4)} & \cellcolor{lightgray!45}24.8 & \cellcolor{lightgray!45}46.0 & \cellcolor{lightgray!45}55.4 & \cellcolor{lightgray!45}57.9(+2.8) & \cellcolor{lightgray!45}38.9 & \cellcolor{lightgray!45}62.0 & \cellcolor{lightgray!45}73.7\\
    % \cline{2-10}
    % &\gc{CWD\dag \cite{shu2021cwd}} & \gc{42.0(+3.4)} &\gc{24.1} &\gc{46.1} &\gc{55.0} & \gc{-}&\gc{-}&\gc{-}&\gc{-} \\
    % &\gc{FGD\dag \cite{yang2022fgd}} & \gc{42.0(+3.4)}&\gc{24.0}&\gc{45.7}&\gc{55.6}&\gc{58.2(+3.1)}&\gc{37.8}&\gc{62.2}&\gc{73.3}\\
    % &\gc{MGD\dag \cite{yang2022mgd}} &\gc{42.3(+3.7)} &\gc{24.4}& \gc{46.2} &\gc{55.9} & \gc{58.4(+3.3)} & \gc{40.4} & \gc{62.3} & \gc{73.9}\\
    \midrule
    \midrule
    \multirow{2}{*}{Teacher} &
    \multirow{2}{*}{Student} &
    \multicolumn{4}{c|}{Boundingbox AP}&
    \multicolumn{4}{c}{Mask AP} \\
    % \cmidrule{3-10}
    & &mAP &AP$_{S}$ & AP$_{M}$ & AP$_{L}$& mAP&  AP$_{S}$ &  AP$_{M}$ & AP$_{L}$\\
    \midrule
    \multirow{5}{*}{\makecell{Cascade\\Mask RCNN\\ResNeXt101}}
    &Mask RCNN-ResNet50 & 39.2&22.9&42.6&51.2&35.4&19.1&38.6&48.4\\
    &FKD\cite{zhang2020fkd} & 41.7(+2.5)&23.4&45.3&55.8&37.4(+2.0)&19.7&40.5&52.1\\
    &FGD\cite{yang2022fgd}& 42.1(+2.9)&23.7&46.2&55.7&37.8(+2.4)&19.7&41.3&52.3\\
    &MGD\cite{yang2022mgd} &42.3(+3.1)&23.9&46.3&56.2&38.1(+2.7) &17.1 &41.1 &56.3\\
    &\cellcolor{lightgray!45}Ours & \cellcolor{lightgray!45}42.4(+3.2) & \cellcolor{lightgray!45}23.8 & \cellcolor{lightgray!45}46.3 & \cellcolor{lightgray!45}56.6 & \cellcolor{lightgray!45}38.2(+2.8) & \cellcolor{lightgray!45}17.3 & \cellcolor{lightgray!45}41.2 & \cellcolor{lightgray!45}56.6\\
    \bottomrule
  \end{tabular}
  % \vspace{-2mm}
  \caption{Results of detectors on COCO dataset. 
     % \dag The method which uses inheriting strategy is reported for reference (\gc{gray}).
%   \dag\ means using inheriting strategy, which can only be applied when the student and teacher have the same head structure.
  }
  \label{table:more results}
\end{table*}

  We compare our method with previous state-of-the-art methods designed for object detection~\cite{zhang2020fkd,yang2022fgd} and a recent generic distillation method~\cite{yang2022mgd}. As shown in Table~\ref{table:more results}, our simple method can achieve competitive results. For example, on the two-stage detector Faster RCNN-ResNet50, we get the mAP of the student model to rise from 38.4 to 42.3, surpassing the previous state-of-the-art method. On the anchor-based single-stage detector RetinaNet-ResNet50 and the anchor-free single-stage detector Reppoints-ResNet50, we also achieve mAP increases of 3.6 and 3.4, respectively, which are comparable to the results of the state-of-the-art method.

\paragraph{Instance Segmentation}
Our method demonstrates its effectiveness on the instance segmentation task, as shown in Table~\ref{table:more results}. The results show that our simple approach leads to +3.2\% improvement in bounding box AP and +2.4\% improvement in mask AP, respectively, outperforming state-of-the-art methods. 
\label{sec:det_experiment}

\subsection{Semantic segmentation on CityScapes}
\paragraph{Settings} For the semantic segmentation task, we evaluate our method with the CityScapes dataset\cite{cordts2016cityscapes}. Specifically, our experiments are conducted on 2975 training images and 500 validation images, and the evaluation metric is mIoU. The models are trained for 40,000 iterations using the SGD optimizer on 8 GPUs with a batch size of 2. 

We conduct experiments on two model configurations: a) a homogeneous setting with PSPNet-Res101 as the teacher model and PSPNet-Res18 as the student model, and b) a heterogeneous setting with PSPNet-Res101 as the teacher model and DeepLabv3-Res18 as the student model. The input size for both configurations is set to $512\times512$, and the distillation loss is computed from the features of the last stage of the model neck. The distillation loss weight $\alpha$ for the homogeneous set is set to $2\times10^{-5}$, and for the heterogeneous set is $\alpha$ to $1\times10^{-5}$.

\paragraph{Results} 

As shown in Table~\ref{table:segmentation results}, our method achieves remarkable results in both homogeneous and heterogeneous configurations. Specifically, the ResNet-18-based PspNet model obtains a mIoU increase of +4.66\% under the homogeneous setting, and the ResNet-18-based deeplabv3 model obtains a mIoU increase of +3.28\% under the heterogeneous setting. Furthermore, when compared to the state-of-the-art method MGD\cite{yang2022mgd}, our method achieves an improvement of +0.88\% mIoU and +0.53\% mIoU on the homogeneous and heterogeneous settings, respectively. These results demonstrate the effectiveness of our method on the semantic segmentation task.
\label{sec:seg_experiment}
\subsection{Ablation Studies and Analysis}

\subsubsection{Benefits of Channel-wise Transformation}

As shown in Table~\ref{table:DifferentTasks}, directly using the features from the teacher and the student without channel-wise transformation can result in significant distillation performance drops in the semantic segmentation task.

To better understand this phenomenon, we calculate the $L_2$-distance between the student feature map and the teacher feature map on the validation dataset. 

The results in Table~\ref{table:L2Dis} show that directly mimicking the teacher feature (corresponding to ‘Identity’ transformation) can achieve a lower $L_2$-distance to the teacher, but obtain significantly poorer performance compared to those using channel-wise transformations. Compared with it, channel-wise transformation methods can obtain an even lower $L_2$-distance after the channel-wise transformation, but the $L_2$-distance before the channel-wise transformation is much larger. Moreover, the distillation performance of the channel-wise transformation methods is much better than directly mimicking.

In the process of distillation, the student model is supervised by two signals: distillation losses and task-specific losses. We conjecture that the limited capacity of the student model makes it difficult to fully capture the knowledge of the teacher, and applying strict distillation constraints (\textit{i.e}., directly mimicking the teacher feature) may over-optimize the student feature with the distillation supervision and prevent them from being trained with the task-specific supervision, leading to performance degradation. On the contrary, our method exploits the channel-wise transformation module to achieve a better balance between the task-specific supervision and the distillation supervision.
%better balancing the student features with the distillation supervision.
\begin{table}[t]
% \small
\centering
\begin{tabular}{lccc}
\toprule
\multirow{2}{*}{\makecell{Transformation}} & \multicolumn{2}{c}{$L_2$-distance} \\
\cmidrule(lr){2-3}
& Before & After & mIoU\\
\midrule
Identity & 0.217 & 0.217 & 46.2\\
Linear & 0.4269 & 0.037 & 71.4\\
MLP(ours) & 0.7691 & 0.032 & 73.5\\
\bottomrule
\end{tabular}
% \vspace{-2mm}
\caption{$L_2$ distances with teacher feature and mIoU scores for different transformations in the semantic segmentation task.}
\label{table:L2Dis}
\end{table}

\subsubsection{Ablation of Transformation Modules}

In this section, we further demonstrate the importance of the channel-wise transformation module in cases where the size of the teacher's feature and the student's feature are not equivalent, \textit{i.e.}, when the number of channels is unequal. We show this by performing distillation of PspNet-Res101 onto DeepLabV3-Res18 on the CityScapes dataset. 

As shown in Table~\ref{table:Transform}, the student model only achieves minimal improvement when there is only a single linear layer without non-linear activation (\textit{i.e.}, ReLU). These results demonstrate that the non-linear transformation plays an important role in improving the student's representation ability.  Compared with MLP, additional local spatial transformation with non-linear activation (implemented with Conv3$\times$3-ReLU-Conv3$\times$3) achieves worse performance. Besides, global spatial transformation with Non-Local block~\cite{wang2018non}  achieves the lowest mIoU. 

These results show that the transformation of spatial dimension does not bring additional gain to our method. We conjecture that an overly complex and powerful learnable transformation will make the distillation process concentrate on optimizing the transformation module rather than the student network itself.
\begin{table}[t]
\small
    \centering

    \begin{tabular}{cccc|c}
    \toprule
\multicolumn{4}{c|}{Transform}            & \multirow{2}{*}{mIoU} \\
Module  & Channel & Spatial & Non-Linear &                       \\
\midrule
Stu-Baseline  & -     & - & -    & 73.20                  \\
Linear  & \cmark     & \xmarkg & \xmarkg    & 73.40                  \\
MLP     & \cmark     & \xmarkg & \cmark        & 76.55                 \\
Conv3$\times$3 & \cmark     & Local     & \cmark        & 75.92              \\
Non-Local~\cite{wang2018non} & \cmark     & Global    & \cmark        & 72.05              \\
    \bottomrule
  \end{tabular}
  % \vspace{-2mm}
  \caption{Performance comparison of different transform modules on semantic segmentation task. The results indicate that our proposed channel-wise non-linear transformation module (MLP) outperforms other methods.}
  \label{table:Transform}
\end{table}

%   \begin{tabular}{lc}
    % \toprule
    % Transform modules  & Linear & MLP &  \\
    % mIoU & 73.40 & 76.55 & 75.92 \\
    %  Transform modules & mIoU  \\
    %  \midrule
    %  Linear (Conv1x1) & 73.40 \\
    %  MLP (Conv1x1-ReLU-Conv1x1) & 76.55 \\
    %  Conv3x3-ReLU-Conv3x3 & 75.92 \\ 
    
    %      Transform modules & mIoU  \\
    %  \midrule
    %  Linear (Conv1x1) & 73.40 \\
    %  MLP (Conv1x1-ReLU-Conv1x1) & 76.55 \\
    %  Conv3x3-ReLU-Conv3x3 & 75.92 \\ 
    %  Attention & - & - \\

%MLP\_double     & \cmark     & \xmarkg & \cmark        & 76.10                 \\
%MLP\_half     & \cmark     & \xmarkg & \cmark        & 76.63  \\
%MLP\_quarter     & \cmark     & \xmarkg & \cmark        & 76.4  \\
%MLP\_half-quarter & \cmark     & \xmarkg & \cmark        & 76.75  \\

\subsubsection{Location of MLP}
Previous research on feature-based distillation techniques, such as FKD~\cite{zhang2020fkd} and FGD~\cite{yang2022fgd}, have employed complex masks in their transformation modules to transform both the student and teacher features. In contrast, our method utilizes a learnable Multi-Layer Perceptron (MLP) for feature transformation. 

\begin{figure}
    \centering
    \includegraphics[width = 0.5\textwidth]{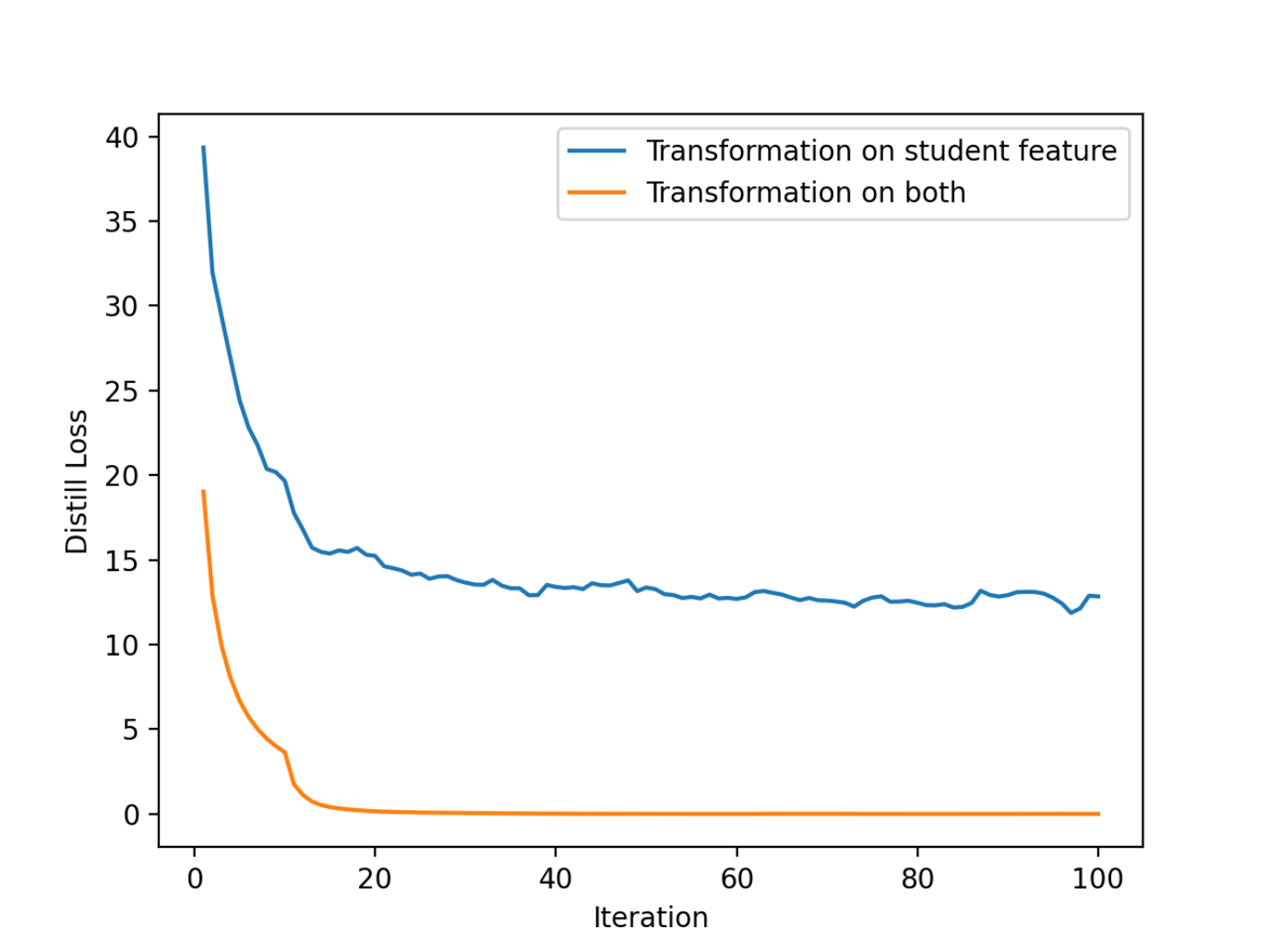}
    % \vspace{-4mm}
    \caption{Comparison of the distillation loss when transforming both the student and teacher features versus transforming only the student features.}
    % \vspace{-2mm}
    \label{fig:loss_comp}
\end{figure}

 The advantage of using a learnable MLP is that it can guide the student to better learn the representations from the teacher. However, applying the transformation to both the student and teacher features can quickly lead to a trivial solution where the distillation loss approaches zero, as demonstrated in Fig.~\ref{fig:loss_comp}. This invalidates the effectiveness of feature distillation. To avoid this problem, our method only applies the transformation to the student feature.

\label{sec:experiment}
\section{Conclusion}

In this paper, we first present a novel discovery that aligning the feature maps between teacher and student along the channel-wise dimension is also effective for addressing the feature misalignment issue in feature-based knowledge distillation. Then, we exploit a Multi-Layer Perceptron (MLP) as the channel-wise transformation module to align the features of the student and the teacher model. Further, we propose a simple and generic framework for feature distillation based on it, with only one hyper-parameter to balance the distillation loss and the task specific loss. Extensive experimental are conducted and the results demonstrate that the proposed method can achieves significant performance improvements in various computer vision tasks including image classification, object detection, instance segmentation, and semantic segmentation, even outperforming the state-of-the-art feature-based knowledge distillation methods in some tasks.

\label{sec:conclusion}

{\small
\bibliographystyle{ieee_fullname}
\bibliography{main}

\begin{thebibliography}{10}\itemsep=-1pt

\bibitem{cao2019gcnet}
Yue Cao, Jiarui Xu, Stephen Lin, Fangyun Wei, and Han Hu.
\newblock Gcnet: Non-local networks meet squeeze-excitation networks and
  beyond.
\newblock In {\em Proceedings of the IEEE/CVF international conference on
  computer vision workshops}, pages 0--0, 2019.

\bibitem{reviewkd}
Pengguang Chen, Shu Liu, Hengshuang Zhao, and Jiaya Jia.
\newblock Distilling knowledge via knowledge review.
\newblock In {\em CVPR}, 2021.

\bibitem{2021mmrazor}
MMRazor Contributors.
\newblock Openmmlab model compression toolbox and benchmark.
\newblock \url{https://github.com/open-mmlab/mmrazor}, 2021.

\bibitem{cordts2016cityscapes}
Marius Cordts, Mohamed Omran, Sebastian Ramos, Timo Rehfeld, Markus Enzweiler,
  Rodrigo Benenson, Uwe Franke, Stefan Roth, and Bernt Schiele.
\newblock The cityscapes dataset for semantic urban scene understanding.
\newblock In {\em Proceedings of the IEEE conference on computer vision and
  pattern recognition}, pages 3213--3223, 2016.

\bibitem{cybenko1989approximation}
George Cybenko.
\newblock Approximation by superpositions of a sigmoidal function.
\newblock {\em Mathematics of control, signals and systems}, 2(4):303--314,
  1989.

\bibitem{dai2021GID}
Xing Dai, Zeren Jiang, Zhao Wu, Yiping Bao, Zhicheng Wang, Si Liu, and Erjin
  Zhou.
\newblock General instance distillation for object detection.
\newblock In {\em Proceedings of the IEEE/CVF Conference on Computer Vision and
  Pattern Recognition}, pages 7842--7851, 2021.

\bibitem{deng2009imagenet}
Jia Deng, Wei Dong, Richard Socher, Li-Jia Li, Kai Li, and Li Fei-Fei.
\newblock Imagenet: A large-scale hierarchical image database.
\newblock In {\em 2009 IEEE conference on computer vision and pattern
  recognition}, pages 248--255. Ieee, 2009.

\bibitem{gou2021knowledge}
Jianping Gou, Baosheng Yu, Stephen~J Maybank, and Dacheng Tao.
\newblock Knowledge distillation: A survey.
\newblock {\em International Journal of Computer Vision}, 129(6):1789--1819,
  2021.

\bibitem{guo2021defeat}
Jianyuan Guo, Kai Han, Yunhe Wang, Han Wu, Xinghao Chen, Chunjing Xu, and Chang
  Xu.
\newblock Distilling object detectors via decoupled features.
\newblock In {\em Proceedings of the IEEE/CVF Conference on Computer Vision and
  Pattern Recognition}, pages 2154--2164, 2021.

\bibitem{resnet}
Kaiming He, Xiangyu Zhang, Shaoqing Ren, and Jian Sun.
\newblock Deep residual learning for image recognition.
\newblock In {\em Proceedings of the IEEE conference on computer vision and
  pattern recognition}, pages 770--778, 2016.

\bibitem{heo2019overhaul}
Byeongho Heo, Jeesoo Kim, Sangdoo Yun, Hyojin Park, Nojun Kwak, and Jin~Young
  Choi.
\newblock A comprehensive overhaul of feature distillation.
\newblock In {\em Proceedings of the IEEE/CVF International Conference on
  Computer Vision}, pages 1921--1930, 2019.

\bibitem{hinton2015distilling}
Geoffrey Hinton, Oriol Vinyals, and Jeffrey Dean.
\newblock Distilling the knowledge in a neural network.
\newblock In {\em NIPS Deep Learning and Representation Learning Workshop},
  2015.

\bibitem{jang2022glamd}
Younho Jang, Wheemyung Shin, Jinbeom Kim, Simon Woo, and Sung-Ho Bae.
\newblock Glamd: Global and local attention mask distillation for object
  detectors.
\newblock In {\em European Conference on Computer Vision}, pages 460--476.
  Springer, 2022.

\bibitem{kzagoruyko2016at}
Nikos Komodakis and Sergey Zagoruyko.
\newblock Paying more attention to attention: improving the performance of
  convolutional neural networks via attention transfer.
\newblock In {\em ICLR}, 2017.

\bibitem{li2022rm}
Gang Li, Xiang Li, Yujie Wang, Shanshan Zhang, Yichao Wu, and Ding Liang.
\newblock Knowledge distillation for object detection via rank mimicking and
  prediction-guided feature imitation.
\newblock In {\em Proceedings of the AAAI Conference on Artificial
  Intelligence}, volume~36, pages 1306--1313, 2022.

\bibitem{lin2022tat}
Sihao Lin, Hongwei Xie, Bing Wang, Kaicheng Yu, Xiaojun Chang, Xiaodan Liang,
  and Gang Wang.
\newblock Knowledge distillation via the target-aware transformer.
\newblock In {\em Proceedings of the IEEE/CVF Conference on Computer Vision and
  Pattern Recognition}, pages 10915--10924, 2022.

\bibitem{lin2017retina}
Tsung-Yi Lin, Priya Goyal, Ross Girshick, Kaiming He, and Piotr Doll{\'a}r.
\newblock Focal loss for dense object detection.
\newblock In {\em Proceedings of the IEEE international conference on computer
  vision}, pages 2980--2988, 2017.

\bibitem{lin2014coco}
Tsung-Yi Lin, Michael Maire, Serge Belongie, James Hays, Pietro Perona, Deva
  Ramanan, Piotr Doll{\'a}r, and C~Lawrence Zitnick.
\newblock Microsoft coco: Common objects in context.
\newblock In {\em European conference on computer vision}, pages 740--755.
  Springer, 2014.

\bibitem{liu2019skds}
Yifan Liu, Ke Chen, Chris Liu, Zengchang Qin, Zhenbo Luo, and Jingdong Wang.
\newblock Structured knowledge distillation for semantic segmentation.
\newblock In {\em Proceedings of the IEEE/CVF Conference on Computer Vision and
  Pattern Recognition}, pages 2604--2613, 2019.

\bibitem{rkd}
Wonpyo Park, Dongju Kim, Yan Lu, and Minsu Cho.
\newblock Relational knowledge distillation.
\newblock In {\em CVPR}, 2019.

\bibitem{park2019rkd}
Wonpyo Park, Dongju Kim, Yan Lu, and Minsu Cho.
\newblock Relational knowledge distillation.
\newblock In {\em Proceedings of the IEEE/CVF Conference on Computer Vision and
  Pattern Recognition}, pages 3967--3976, 2019.

\bibitem{ren2015faster}
Shaoqing Ren, Kaiming He, Ross Girshick, and Jian Sun.
\newblock Faster r-cnn: Towards real-time object detection with region proposal
  networks.
\newblock {\em Advances in neural information processing systems}, 28, 2015.

\bibitem{romero2014fitnets}
Adriana Romero, Nicolas Ballas, Samira~Ebrahimi Kahou, Antoine Chassang, Carlo
  Gatta, and Yoshua Bengio.
\newblock Fitnets: Hints for thin deep nets.
\newblock {\em arXiv preprint arXiv:1412.6550}, 2014.

\bibitem{shu2021cwd}
Changyong Shu, Yifan Liu, Jianfei Gao, Zheng Yan, and Chunhua Shen.
\newblock Channel-wise knowledge distillation for dense prediction.
\newblock In {\em Proceedings of the IEEE/CVF International Conference on
  Computer Vision}, pages 5311--5320, 2021.

\bibitem{tian2019crd}
Yonglong Tian, Dilip Krishnan, and Phillip Isola.
\newblock Contrastive representation distillation.
\newblock In {\em International Conference on Learning Representations}, 2019.

\bibitem{crd}
Yonglong Tian, Dilip Krishnan, and Phillip Isola.
\newblock Contrastive representation distillation.
\newblock In {\em ICLR}, 2020.

\bibitem{wang2019fgfi}
Tao Wang, Li Yuan, Xiaopeng Zhang, and Jiashi Feng.
\newblock Distilling object detectors with fine-grained feature imitation.
\newblock In {\em Proceedings of the IEEE/CVF Conference on Computer Vision and
  Pattern Recognition}, pages 4933--4942, 2019.

\bibitem{wang2018non}
Xiaolong Wang, Ross Girshick, Abhinav Gupta, and Kaiming He.
\newblock Non-local neural networks.
\newblock In {\em Proceedings of the IEEE conference on computer vision and
  pattern recognition}, pages 7794--7803, 2018.

\bibitem{wang2020ifvd}
Yukang Wang, Wei Zhou, Tao Jiang, Xiang Bai, and Yongchao Xu.
\newblock Intra-class feature variation distillation for semantic segmentation.
\newblock In {\em European Conference on Computer Vision}, pages 346--362.
  Springer, 2020.

\bibitem{Xie2016resnext}
Saining Xie, Ross Girshick, Piotr Doll{\'a}r, Zhuowen Tu, and Kaiming He.
\newblock Aggregated residual transformations for deep neural networks.
\newblock In {\em Proceedings of the IEEE conference on computer vision and
  pattern recognition}, pages 1492--1500, 2017.

\bibitem{yang2022cirkd}
Chuanguang Yang, Helong Zhou, Zhulin An, Xue Jiang, Yongjun Xu, and Qian Zhang.
\newblock Cross-image relational knowledge distillation for semantic
  segmentation.
\newblock In {\em Proceedings of the IEEE/CVF Conference on Computer Vision and
  Pattern Recognition}, pages 12319--12328, 2022.

\bibitem{yang2022fgd}
Zhendong Yang, Zhe Li, Xiaohu Jiang, Yuan Gong, Zehuan Yuan, Danpei Zhao, and
  Chun Yuan.
\newblock Focal and global knowledge distillation for detectors.
\newblock In {\em Proceedings of the IEEE/CVF Conference on Computer Vision and
  Pattern Recognition}, pages 4643--4652, 2022.

\bibitem{yang2022mgd}
Zhendong Yang, Zhe Li, Mingqi Shao, Dachuan Shi, Zehuan Yuan, and Chun Yuan.
\newblock Masked generative distillation.
\newblock {\em ECCV}, 2022.

\bibitem{zhang2020fkd}
Linfeng Zhang and Kaisheng Ma.
\newblock Improve object detection with feature-based knowledge distillation:
  Towards accurate and efficient detectors.
\newblock In {\em International Conference on Learning Representations}, 2020.

\bibitem{zhao2022dkd}
Borui Zhao, Quan Cui, Renjie Song, Yiyu Qiu, and Jiajun Liang.
\newblock Decoupled knowledge distillation.
\newblock In {\em Proceedings of the IEEE/CVF Conference on computer vision and
  pattern recognition}, pages 11953--11962, 2022.

\bibitem{zhao2017pspnet}
Hengshuang Zhao, Jianping Shi, Xiaojuan Qi, Xiaogang Wang, and Jiaya Jia.
\newblock Pyramid scene parsing network.
\newblock In {\em Proceedings of the IEEE conference on computer vision and
  pattern recognition}, pages 2881--2890, 2017.

\bibitem{zheng2022ld}
Zhaohui Zheng, Rongguang Ye, Ping Wang, Dongwei Ren, Wangmeng Zuo, Qibin Hou,
  and Ming-Ming Cheng.
\newblock Localization distillation for dense object detection.
\newblock In {\em Proceedings of the IEEE/CVF Conference on Computer Vision and
  Pattern Recognition}, pages 9407--9416, 2022.

\bibitem{zhixing2021frs}
Du Zhixing, Rui Zhang, Ming Chang, Shaoli Liu, Tianshi Chen, Yunji Chen, et~al.
\newblock Distilling object detectors with feature richness.
\newblock {\em Advances in Neural Information Processing Systems}, 34, 2021.

\bibitem{zhou2020wsld}
Helong Zhou, Liangchen Song, Jiajie Chen, Ye Zhou, Guoli Wang, Junsong Yuan,
  and Qian Zhang.
\newblock Rethinking soft labels for knowledge distillation: A bias--variance
  tradeoff perspective.
\newblock In {\em International Conference on Learning Representations}, 2020.

\end{thebibliography}
}

\ifarxiv \clearpage \appendix

\section{Additional Experimental Results}

\subsection{Extended Detection Results}
\begin{table*}[t]
  \centering
  \begin{tabular}{l|c|cccccc}
    \toprule
    Method & Schedule & mAP & AP$_{50}$ & AP$_{75}$ &AP$_{S}$& AP$_{M}$ & AR$_{L}$ \\
    \midrule
    %\multirow{7}{*}{\makecell{RetinaNet\\ResNeXt101}}
    RetinaNet-ResNet50(Student) & 1x & 36.5 &55.4&39.1 &20.4&40.3&48.1\\
    RetinaNet-ResNext101(Teacher) &3x & 41.6 &61.4&44.3 &23.9&45.5&54.5\\
    Hint-learning\cite{romero2014fitnets} &1x & 37.1 & 56.5 & 39.2 & 21.4 & 40.7 & 48.8 \\
    FGFI\cite{wang2019fgfi} &1x& 38.4 & 57.5 & 41.1 & 20.8 & 42.0 & 51.9 \\
    FKD\cite{zhang2020fkd} &1x& 39.0 & 58.1 & 41.8 & 22.3 & 42.9 & 51.7 \\
    FRS\cite{zhixing2021frs} &1x& 39.3 & 58.7 & 41.9 & 21.4 & 43.1 & 52.3 \\
    GLAMD\cite{jang2022glamd} &1x& 40.0 & 59.5 & 42.5 & 22.8 & 44.0 & 53.4 \\
    \cellcolor{lightgray!45}Ours &\cellcolor{lightgray!45}1x& \cellcolor{lightgray!45}{40.0(+3.5)} & \cellcolor{lightgray!45}59.4 & \cellcolor{lightgray!45}42.7 & \cellcolor{lightgray!45}22.5 & \cellcolor{lightgray!45}44.5& \cellcolor{lightgray!45}53.0 \\
    % \cline{2-10}
    % &\gc{CWD\dag \cite{shu2021cwd}} & \gc{40.8(+3.4)} &\gc{22.7} &\gc{44.5} &\gc{55.3} & \gc{-}&\gc{-}&\gc{-}&\gc{-} \\
    % &\gc{FGD\dag\cite{yang2022fgd}} & \gc{40.7(+3.3)}&\gc{22.9}&\gc{45.0}&\gc{54.7}&\gc{56.8(+2.9)}&\gc{36.5}&\gc{61.4}&\gc{72.8}\\
    %  &\gc{MGD\dag \cite{yang2022mgd}} &\gc{41.0(+3.6)} &\gc{23.4} &\gc{45.3}& \gc{55.7} & \gc{57.0(+3.1)} & \gc{37.2} & \gc{61.7} & \gc{72.8}\\
    \midrule
    %\multirow{7}{*}{\makecell{Faster-RCNN\\ResNeXt101}}
    Faster-ResNet50(Student) &1x& 37.4 &58.1&40.4&21.2&41.0&48.1\\
    Faster-ResNext101(Teacher) &3x & 43.1 &63.6&47.2 &26.5&46.9&56.0\\
    Hint-learning\cite{romero2014fitnets} &1x& 38.7 & 59.7 & 41.8 & 23.1 & 42.0 & 50.9 \\
    FGFI\cite{wang2019fgfi} &1x& 39.5 & 59.9 & 43.2 & 21.7 & 43.4 & 53.2 \\
    FKD\cite{zhang2020fkd} &1x& 40.1 & 60.8 & 43.4 & 22.9 & 44.1 & 53.1 \\
    FRS\cite{zhixing2021frs} &1x& 40.3 & 61.8 & 43.9 & 23.3 & 44.3 & 52.4 \\
    GLAMD\cite{jang2022glamd} &1x& 40.8 & 61.4 & 44.3 & 23.2 & 45.0 & 53.2 \\
    \cellcolor{lightgray!45}Ours &\cellcolor{lightgray!45}1x& \cellcolor{lightgray!45}{41.3(+3.9)} & \cellcolor{lightgray!45}61.8 & \cellcolor{lightgray!45}44.8 & \cellcolor{lightgray!45}23.7 & \cellcolor{lightgray!45}45.5 & \cellcolor{lightgray!45}54.4 \\
    \midrule
    %\multirow{7}{*}{\makecell{Faster-RCNN\\ResNeXt101}}
    Cascade-ResNet50(Student) &1x& 40.3 &58.6&44.0&22.5&43.8&52.9\\
    Cascade-ResNext101(Teacher) &3x & 44.5 &63.2&48.5 &25.5&48.1&58.4\\
    Hint-learning\cite{romero2014fitnets} &1x& 40.6 & 59.4 & 44.4 & 22.5 & 48.1 & 58.4 \\
    FGFI\cite{wang2019fgfi} &1x& 41.7 & 60.6 & 45.6 & 23.3 & 45.2 & 55.9\\
    FKD\cite{zhang2020fkd} &1x& 42.4 & 60.9 & 46.2 & 23.4 & 46.2 & 56.1 \\
    FRS\cite{zhixing2021frs} &1x& 42.7 & 61.3 & 46.7 & 24.4 & 46.3 & 56.2 \\
    GLAMD\cite{jang2022glamd} &1x& 43.0 & 61.5 & 46.8 & 24.1 & 47.3 & 56.8 \\
    \cellcolor{lightgray!45}Ours &\cellcolor{lightgray!45}1x& \cellcolor{lightgray!45}{43.5(+3.2)} & \cellcolor{lightgray!45}62.0 & \cellcolor{lightgray!45}47.5 & \cellcolor{lightgray!45}24.2 & \cellcolor{lightgray!45}47.7 & \cellcolor{lightgray!45}57.7 \\
    \bottomrule
  \end{tabular}
  \vspace{-2mm}
  \caption{Results of detectors on COCO dataset.}
  \label{table:more_coco_results}
\end{table*}
Consistent with the experimental setting used in \cite{jang2022glamd}, we evaluate the performance of our proposed method with other knowledge distillation methods, utilizing both single-stage (RetinaNet\cite{lin2017retina}) and two-stage detectors (Faster-RCNN~\cite{ren2015faster}). The hyper-parameter $\alpha$ is assigned values of $2\times 10^{-5}$ and $5\times 10^{-7}$ for the single-stage and two-stage detectors, respectively. The results of this evaluation are detailed in Table~\ref{table:more_coco_results}, which demonstrates that our method outperforms all previous approaches in terms of distillation performance. Notably, our method surpasses both FRS~\cite{zhixing2021frs} and GLAMD~\cite{jang2022glamd}, state-of-the-art distillation methods specifically tailored for object detection tasks. These empirical results demonstrate the simplicity and efficacy of our proposed method.

\subsection{Alternative Loss Functions}
\begin{table}[t]
    \centering
  \begin{tabular}{c|cc}
    \toprule
     different Loss & mIoU   \\
     \midrule
     L2 Loss & 76.55  \\
     KL divergence & 76.25  \\
     CWD Loss\cite{shu2021cwd} & 76.53\\
    \bottomrule
  \end{tabular}
  \caption{Different Loss Function.}
  \vspace{-0.05in}
  \label{table:loss}
\end{table}
In line with the majority of feature-based knowledge distillation approaches, our method employs the $L2$-distance as its loss function. In this subsection, we investigate the impact of various loss functions within the context of the PspNet-Res101 distilling DeepLabV3-Res18 setting applied to the CityScapes dataset. The results are presented in Table~\ref{table:loss}. It can be seen that there is no significant gap of using different loss functions, and $L2$-distance show slightly better result.

\subsection{Sensitivity Analysis}
\begin{figure}
\vspace{-5mm}
    \centering
    \includegraphics[width = 0.46\textwidth]{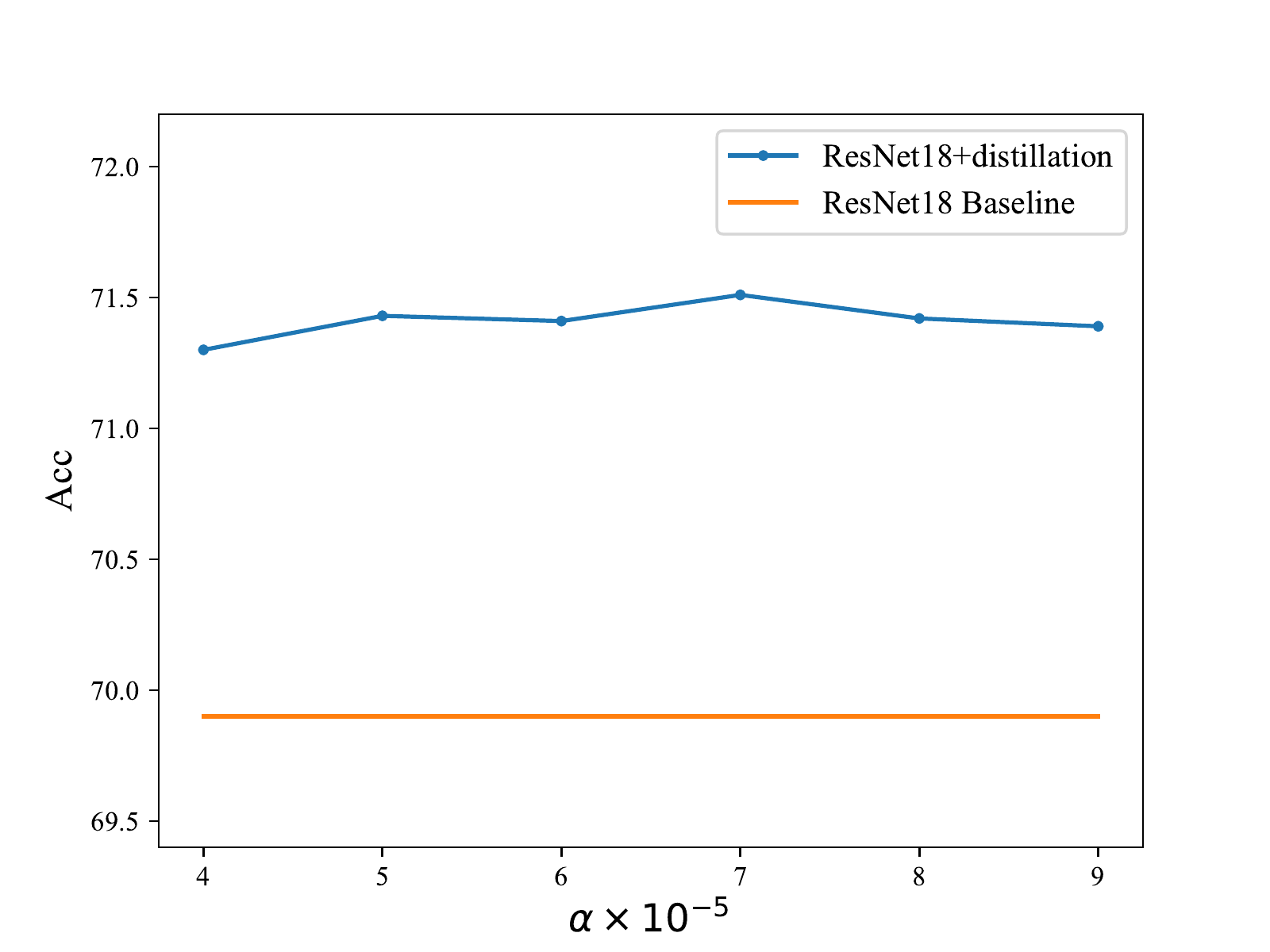}
    \vspace{-3mm}
    \caption{Sensitivity study of hyper-parameters $\alpha$ with ResNet34-ResNet18 on ImageNet classification task.}
    \label{fig:sensitivity}
\end{figure}
\begin{figure}
\vspace{-5mm}
    \centering
    \includegraphics[width = 0.46\textwidth]{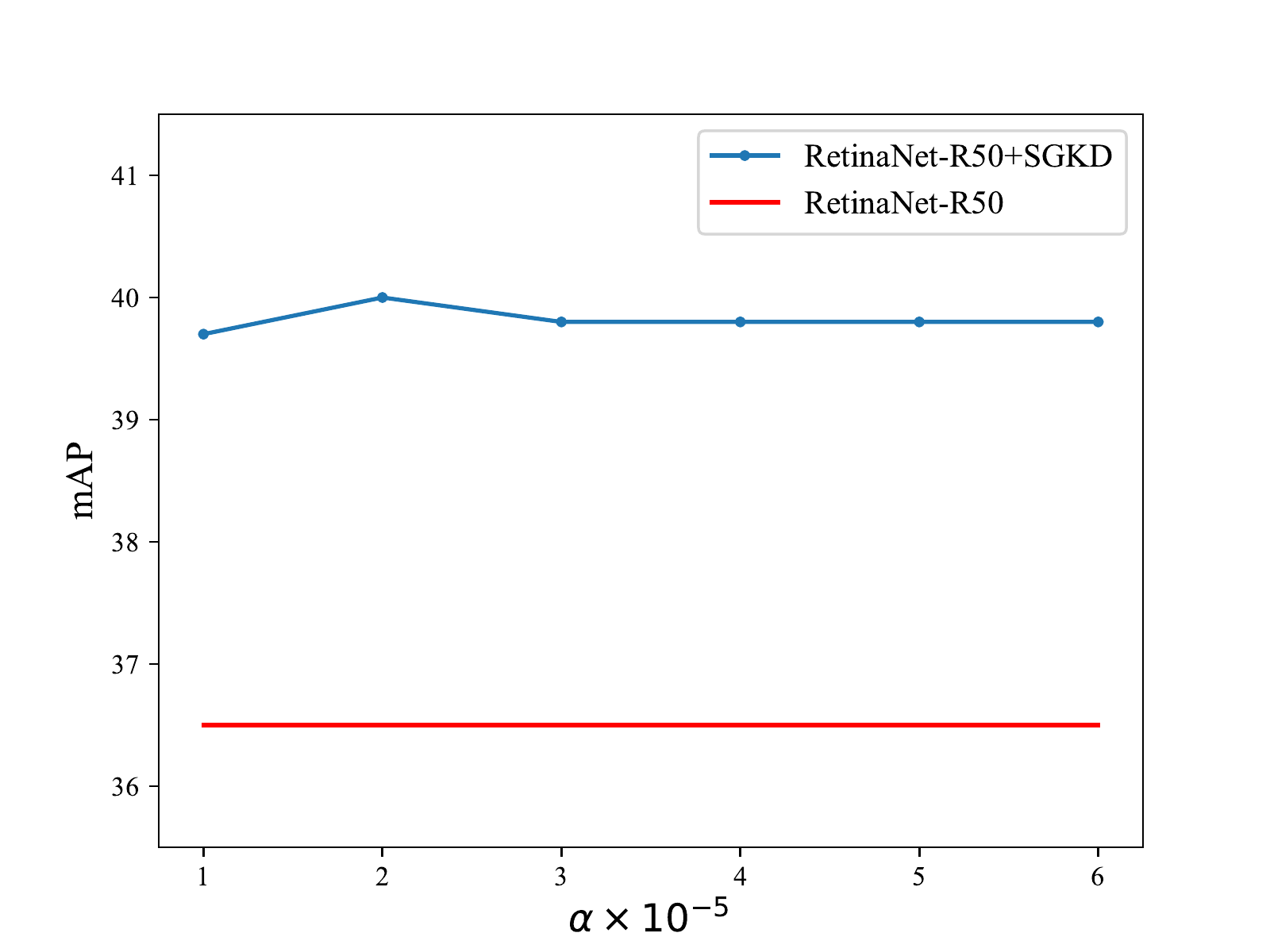}
    \vspace{-3mm}
    \caption{Sensitivity study of hyper-parameters $\alpha$ with RetinaNet-X101 - RetinaNet-R50 on COCO object detection task.}
    \label{fig:more_sensitivity}
\end{figure}
In this subsection, we examine the sensitivity of the hyper-parameter $\alpha$ for both image classification and object detection tasks.

For the image classification task, we use ResNet-34 as the teacher model and ResNet-18 as the student model, employing the ImageNet dataset~\cite{deng2009imagenet}. We specifically train the model on 1.2 million images from the ImageNet training set and test it on 50,000 images from the validation set, using Top-1 accuracy as the evaluation metric. Our training procedure follows a standard approach, consisting of 100 epochs with learning rate decay at the 30th, 60th, and 90th epochs. We utilize Stochastic Gradient Descent (SGD) as the optimizer, with an initial learning rate of 0.1. The training is conducted on 8 GPUs, each with a batch size of 32 images.

For the object detection task, we employ RetinaNet-ResNext101 (3x training schedule) as the teacher model and RetinaNet-ResNet50 as the student model, evaluating our method on the COCO dataset~\cite{lin2014coco}. We train the model on 120,000 images from the COCO training set and test it on 5,000 images from the validation set, using mean Average Precision (mAP) as the evaluation metric. Our training procedure adheres to a standard 1x schedule, encompassing 12 epochs with learning rate reduction at the 8th and 11th epochs. The optimization is carried out using Stochastic Gradient Descent (SGD), and the model is trained on 8 GPUs, each with a batch size of 2.

As illustrated in Fig.~\ref{fig:more_sensitivity}, our method is not sensitive to $\alpha$, which serves only to balance the overall loss.

\section{Additional Visualizations}
\subsection{Impact of the Channel-wise Transformation Module}
\begin{figure}
    \centering
    \includegraphics[width = 0.45\textwidth]{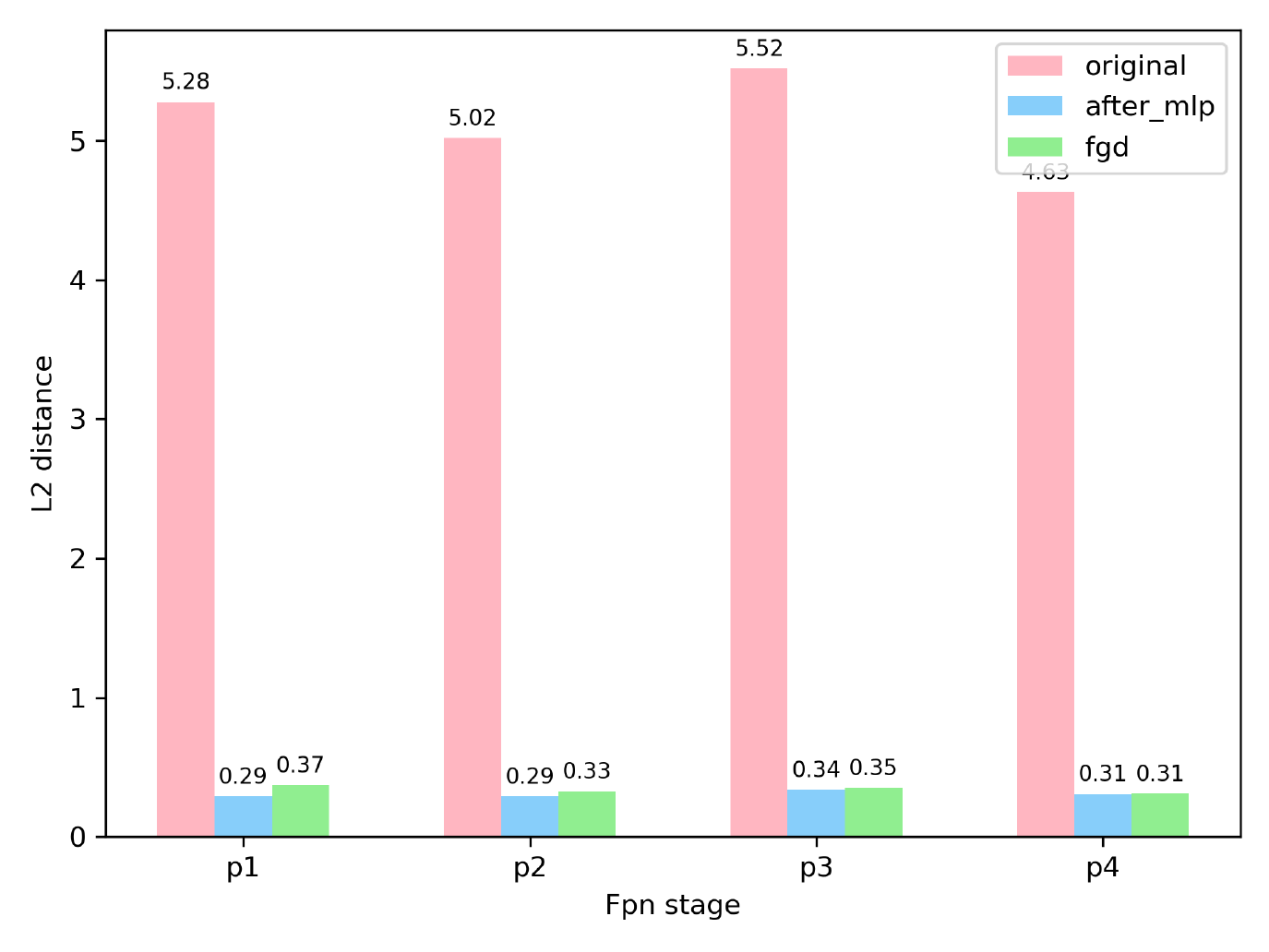}
    \vspace{-4mm}
    \caption{We calculate the $L2$-distance between the student model (Faster RCNN-ResNet50) and the teacher model (CascadeMask RCNN ResNeXt101) on different FPN levels. We calculate the $L2$-distance before and after the channel-wise adaption MLP. The results indicate that the MLP can help narrow the gap between the student and the teacher feature.}    
    \label{fig:l2_after_mlp}
\end{figure}

To further assess the effect of the proposed channel-wise transformation module, we compare our method with FGD~\cite{yang2022fgd} on the detection task. Our method achieves a higher mAP than FGD (42.3 vs. 42.0); however, the student features learned by our method exhibit a larger $L2$-distance with the teacher features compared to the $L2$-distance between the FGD~\cite{yang2022fgd} learned student features and the teacher features. As illustrated in Fig~\ref{fig:l2_after_mlp}, after incorporating the channel-wise transformation module (MLP), the $L2$-distance relative to the teacher features is significantly reduced and becomes lower than the student-teacher feature distance of FGD. These findings suggest that a learnable channel-wise transformation module can act as a semantic translator, facilitating the student model's approximation to the teacher model in an indirect manner.

\subsection{Activation Pattern Visualization}
\begin{figure*}
    \centering
    \includegraphics[width = \textwidth]{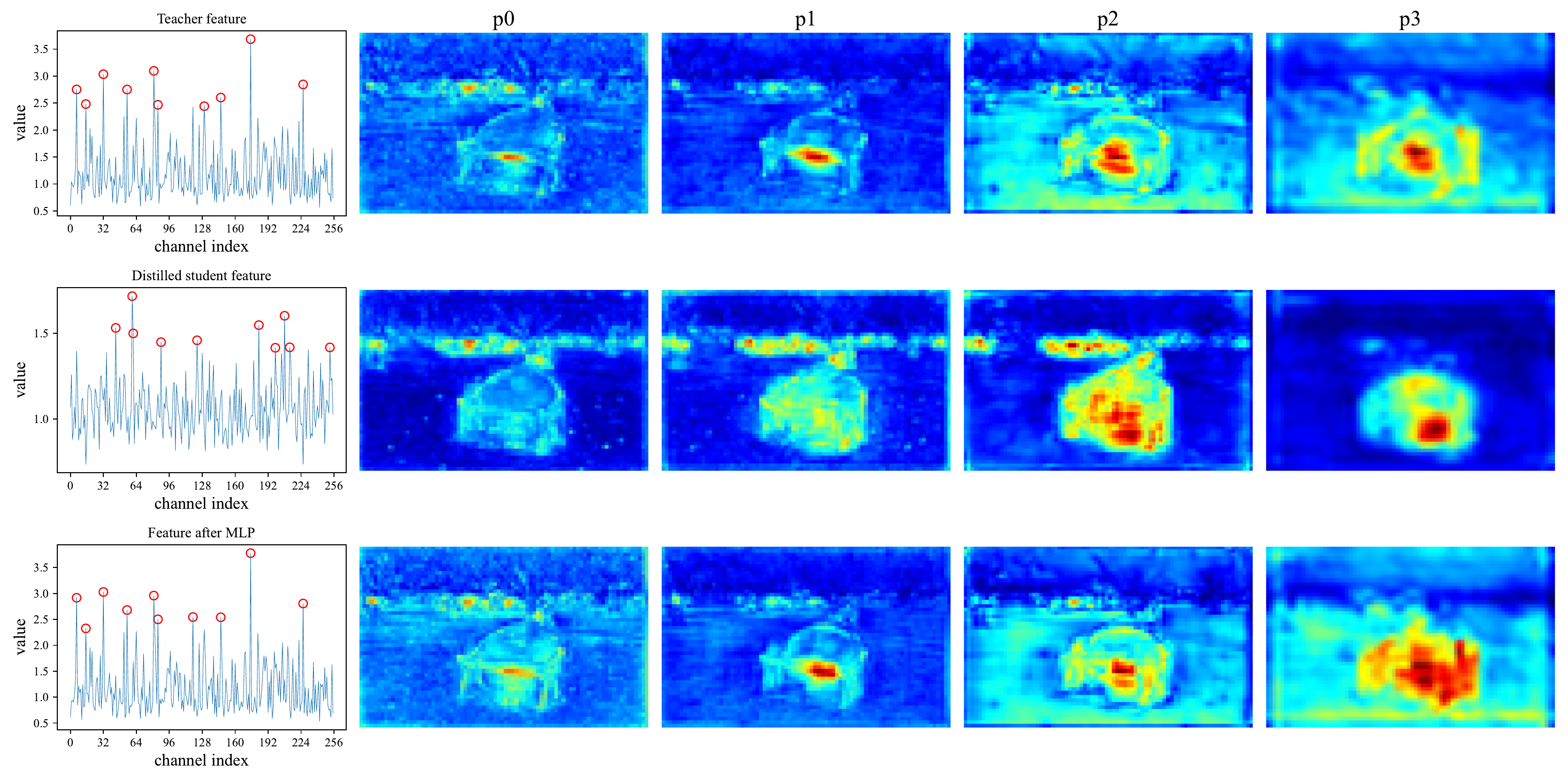}
    \caption{
    An example to illustrate that the MLP module serves as a semantic ``translator''. Left: We adopt a similar way with \cite{yang2022fgd} to calculate the channel-wise attention map, observe the value distribution on different channels, and find that though the student model after distillation shows different distribution to the teacher model, it becomes similar to the channel attention map after the MLP module. Right: Activation patterns of the teacher feature, distilled student feature, and feature after MLP module.
    }
    % \vspace{-2mm}
    \label{fig:visual_activation}
\end{figure*}
In this section, we present an example, depicted in Fig.\ref{fig:visual_activation}, to demonstrate the role of the MLP module as a semantic ``translator''. On the left, we adopt a method akin to that of Yang et al.\cite{yang2022fgd} to calculate the channel-wise attention map, observing the value distribution across different channels. We find that although the student model after distillation exhibits a distinct distribution compared to the teacher model, it becomes more similar to the channel attention map following the implementation of the MLP module. On the right, we display the activation patterns of the teacher feature, distilled student feature, and feature after the MLP module.

\section{Experiment details}
In this section, we present the experiment details in table.\ref{table:DifferentTasks}, 
For the classification task, we use ResNet-34 as the teacher model and ResNet-18 as the student model, conducting feature distillation on the feature map of the last stage of the backbone (both teacher and student feature maps have 512 channels). We employ the ImageNet dataset~\cite{deng2009imagenet}, training the model on 1.2 million images from the ImageNet training set and testing it on 50,000 images from the validation set, using Top-1 accuracy as the evaluation metric. Our training procedure follows a standard approach, consisting of 100 epochs with learning rate decay at the 30th, 60th, and 90th epochs. We utilize Stochastic Gradient Descent (SGD) as the optimizer, with an initial learning rate of 0.1. The training is conducted on 8 GPUs, each with a batch size of 32 images.

For the detection task, we use RetinaNet-ResNext101 as the teacher model and RetinaNet-ResNet50 as the student model. Following FGD~\cite{yang2022fgd}, we conduct distillation on the neck (both teacher and student feature maps have 256 channels). We evaluate our method on the COCO dataset~\cite{lin2014coco}, training the model on 120,000 images from the COCO training set and testing it on 5,000 images from the validation set, using mean Average Precision (mAP) as the evaluation metric. Our training procedure adheres to a standard 1x schedule, encompassing 12 epochs with learning rate reduction at the 8th and 11th epochs. The optimization is carried out using Stochastic Gradient Descent (SGD), and the model is trained on 8 GPUs, each with a batch size of 2.

For the segmentation task, we use PSPNet-ResNet34 as the teacher model and PSPNet-ResNet18 as the student model, conducting feature distillation on the feature map of the last stage of the backbone (both teacher and student feature maps have 512 channels). We evaluate our method using the CityScapes dataset~\cite{cordts2016cityscapes}, conducting experiments on 2,975 training images and 500 validation images, using mean Intersection over Union (mIoU) as the evaluation metric.. The models are trained for 40,000 iterations using the SGD optimizer on 8 GPUs, each with a batch size of 2.

For different Transformation Modules, The "Identity" approach directly mimics features using the l2-distance metric, the "Linear" approach employs a linear conv1x1 transformation, while the "Task-Specific" method utilizes established techniques specifically designed for each task. such as TaT-Cls~\cite{lin2022tat} for classification, which involves using a target-aware transformer. FGD~\cite{yang2022fgd} for object detection, which involves using different attention masks and global context module, and TaT-Seg~\cite{lin2022tat} for semantic segmentation, which involves using patch-group distillation and anchor-point distillation.

 \fi

\end{document}